\documentclass[11pt,english]{article}
\pdfoutput=1

\usepackage{natbib}
\usepackage{graphicx}
\usepackage{bm}
\usepackage[utf8]{inputenc}		
\usepackage{babel}						
\usepackage{xspace}
\usepackage{amsmath}					
\usepackage[frak = euler]{mathalpha}				
\usepackage{algorithm}
\usepackage{multirow}					
\usepackage{colortbl}					
\usepackage{tabularx}
\usepackage[table]{xcolor}		
\usepackage{paralist}					
\usepackage[noend]{algpseudocode}	
\usepackage{stackrel}

\algrenewcommand\algorithmicrequire{\textbf{Voraussetzung:}}
\algrenewcommand\algorithmicensure{\textbf{Abschlussbedingung:}}

\usepackage[scaled]{helvet}							
 
\usepackage[T1]{fontenc}

\usepackage{tikz}
\usetikzlibrary{shapes}
\usetikzlibrary{arrows}				
\usetikzlibrary{positioning}	
\usetikzlibrary{calc}					
\usetikzlibrary{matrix} 	    

\usepackage[labelformat=simple,figurename=Figure, tablename=Table]{caption}	

\newcommand{\latexOrPdflatex}[2]{\ifx\undefined\pdfoutput%
#1%
\else%
#2%
\fi}

\latexOrPdflatex{
\newcommand{\href}[2]{#2}
}{
\usepackage[colorlinks,pagebackref,citecolor=blue,bookmarks=true,pdfstartview=FitH,pdfstartpage=1,pdftitle={General Board Game Playing Framework},pdfauthor={Wolfgang Konen}]{hyperref}     
}

\newcommand{\fcol}{\mbox{\tt fcol}\xspace}
\newcommand{\sloc}{\mbox{\tt sloc}\xspace}
\newcommand{\sL}{\ensuremath{s_{\ell}}\xspace}
\newcommand{\bfTheta}{\mathbf{\Theta}}

\newcommand{\StateObservation}[1]{\href{\#hrefStateObs}{StateObservation}\xspace}
\newcommand{\Sticker}{\href{\#hrefSticker}{sticker}\xspace}
\newcommand{\Stickers}{\href{\#hrefSticker}{stickers}\xspace}
\newcommand{\faceID}{\href{\#hrefFaceID}{face ID}\xspace}
\newcommand{\faceIDs}{\href{\#hrefFaceID}{face IDs}\xspace}

\newcommand{\Cubies}{\href{\#hrefCubie}{cubies}\xspace}
\newcommand{\WholeCubeRotation}{\href{\#hrefWCR}{whole-cube rotation}\xspace}
\newcommand{\WCR}{\href{\#hrefWCR}{WCR}\xspace}
\newcommand{\WCRs}{\href{\#hrefWCR}{WCRs}\xspace}
\newcommand{\HTM}{\href{\#hrefHTM}{HTM}\xspace}
\newcommand{\QTM}{\href{\#hrefQTM}{QTM}\xspace}
\newcommand{\GodsNumber}{\href{\#hrefGodsNum}{God's number}\xspace}
\newcommand{\firstTrafoFirst}{\href{\#hrefFTF}{first-trafo-first}\xspace}
\newcommand{\PlayAgent}[1]{\href{\#hrefPlayAgent}{PlayAgent}\xspace}
\newcommand{\AgentBase}[1]{\href{\#hrefPlayAgent}{AgentBase}\xspace}
\newcommand{\AgentState}[1]{\href{\#hrefAgentState}{AgentState}\xspace}
\newcommand{\NPlayerGames}[1]{\href{\#hrefNPlayer}{$N$-player games}\xspace}
\newcommand{\FinalAdaptStep}[1]{\href{\#hrefFinalAdapt}{final adaptation step}\xspace}
\newcommand{\TCLbase}{\textit{TCL-base}\xspace}
\newcommand{\TCLwrap}{\textit{TCL-wrap}\xspace}
\newcommand{\BoardVector}{\href{\#hrefBV}{BoardVector}\xspace}
\newcommand{\BoardVectors}{\href{\#hrefBV}{BoardVectors}\xspace}
\newcommand{\DAVI}{\href{\#hrefDAVI}{DAVI}\xspace}
\newcommand{\DRB}{\href{\#hrefDRBcubie}{(DRB)}\xspace}

\newcommand{\bl}{\cellcolor[RGB]{196, 188, 252}}  
\newcommand{\co}{\cellcolor[RGB]{248, 222, 170}}	
\newcommand{\gr}{\cellcolor[RGB]{144, 248, 171}}  
\newcommand{\ye}{\cellcolor[RGB]{251, 254, 220}}  
\newcommand{\re}{\cellcolor[RGB]{243, 167, 167}}  
\newcolumntype{a}{>{\columncolor[RGB]{240, 240, 255}}c}   
\newcolumntype{j}{>{\columncolor[RGB]{247, 252, 200}}c}   
\newcolumntype{v}{>{\columncolor[RGB]{251, 254, 220}}c}   
\newcolumntype{x}{>{\columncolor[RGB]{196, 188, 252}}c}   
\newcolumntype{y}{>{\columncolor[RGB]{144, 248, 171}}c}		
\newcolumntype{z}{>{\columncolor[RGB]{243, 167, 167}}c}		

\title{%
\vspace{-0.5cm}{\small \raggedleft{e-print 
\href{http://www.gm.fh-koeln.de/ciopwebpub/Konen22b.d/TR-Rubiks.pdf}
		 {\url{http://www.gm.fh-koeln.de/ciopwebpub/Konen22b.d/TR-Rubiks.pdf}}}} \\
\vspace{1.cm} 
Towards Learning Rubik’s Cube with \\
N-tuple-based Reinforcement Learning\\
}
\author{Wolfgang Konen \\ \\
 Technical Report, \\
 Computer Science Institute, \\
 TH Köln, \\
 University of Applied Sciences, \\
 Germany \\[0.3cm]
 \texttt{\href{mailto:wolfgang.konen@th-koeln.de}{wolfgang.konen@th-koeln.de}} \\[0.3cm]
 Sep 2022,\\
 last update Jan 2023
}

\date{}

\begin{document} 

\maketitle 

\begin{abstract}
This work describes in detail how to learn and solve the Rubik's cube game (or puzzle) in the General Board Game (GBG) learning and playing framework. We cover the  cube sizes 2x2x2 and 3x3x3. We describe in detail the cube's state representation, how to transform it with twists, whole-cube rotations and color transformations and explain the use of symmetries in Rubik's cube. Next, we discuss different n-tuple representations for the cube, how we train the agents by reinforcement learning and how we improve the trained agents during evaluation by MCTS wrapping.

We present results for agents that learn Rubik's cube from scratch, with and without MCTS wrapping, with and without symmetries and show that both, MCTS wrapping and symmetries, increase computational costs, but lead at the same time to much better results. We can solve the 2x2x2 cube completely, and the 3x3x3 cube in the majority of the cases for scrambled cubes up to $p=15$ (QTM). We cannot yet reliably solve 3x3x3 cubes with more than 15 scrambling twists.

Although our computational costs are higher with MCTS wrapping and with symmetries than without, they are still considerably lower than in the approaches of \cite{mcaleer2018solving,mcaleer2019solving} and \cite{agostinelli2019solving} who provide the best Rubik's cube learning agents so far.
\end{abstract}

\newpage
\tableofcontents

\newpage
\section{Introduction} 
\label{sec:introduction}

\subsection{Motivation}
\label{sec:motivation}

Game learning and game playing is an  interesting test bed for strategic decision making. Games usually have large state spaces, and they often require complex pattern recognition and strategic planning capabilities to decide which move is the best in a certain situation. If algorithms learn a game (or, even better, a variety of different games) just by self-play, given no other knowledge than the game rules, it is likely that they perform also well on other problems of strategic decision making. 

In recent years, reinforcement learning (RL) and deep neural networks (DNN) achieved superhuman capabilities in a number of competitive games \citep{mnih2015human,silver2016AlphaGo}. This success has been a product of the combination of reinforcement learning, deep learning and Monte Carlo Tree Search (MCTS). However, current deep reinforcement learning (DRL) methods struggle in environments with a high number of states and a small number of reward states. 

\begin{figure}[h]
\centerline{
    \begin{tabular}{cc}
    \includegraphics[width=0.28\columnwidth]{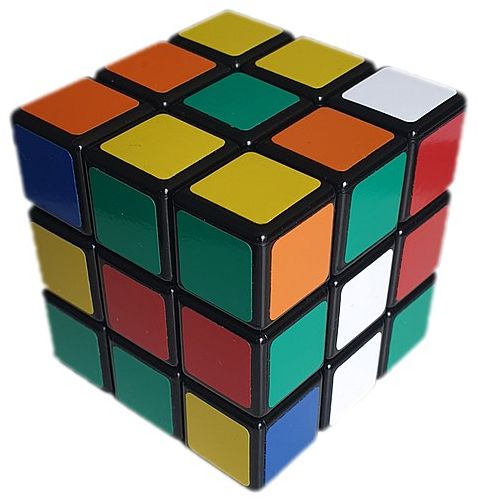}
             &  \hspace{0.05\columnwidth}
    \includegraphics[width=0.19\columnwidth,height=0.19\columnwidth]{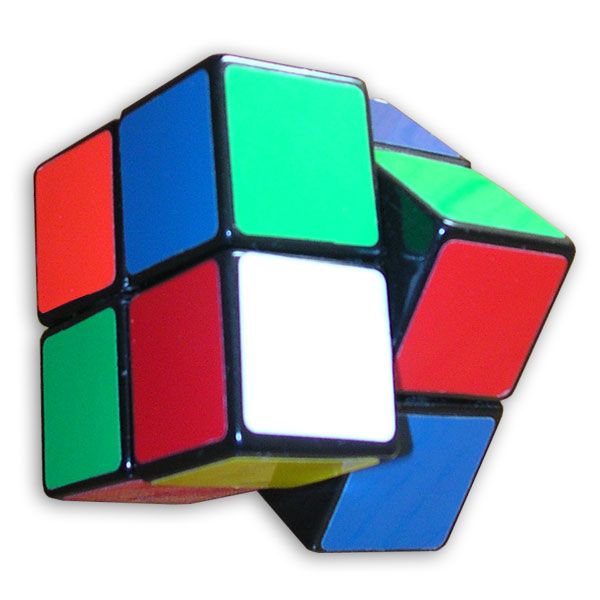}
         \\
 (a)     &    \hspace{0.05\columnwidth} (b) 
    \end{tabular}
} 
\caption{(a) Scrambled 3x3x3 Rubik's Cube. (b) 2x2x2 cube in the middle of a twist.
}
\label{fig:gamesRubik}
\end{figure}

The Rubik's cube puzzle is an example of such an environment since the classical 3x3x3 cube has $4.3\cdot 10^{19}$ states and only \textit{one} state (the solved cube) has a reward. A somewhat simpler puzzle is the 2x2x2 cube with $3.6\cdot 10^6$ state and again only one reward state. Both cubes are shown in Fig.~\ref{fig:gamesRubik}. 

The difficult task to \textit{learn from scratch} how to solve arbitrary scrambled cubes (i.e. without being taught by expert knowledge, whether from humans or from computerized solvers) was not achievable with DRL methods for a long time. Recently, the works of \cite{mcaleer2018solving,mcaleer2019solving} and \cite{agostinelli2019solving} provided a breakthrough in that direction (see Sec.~\ref{sec:McAleer} and \ref{sec:rel-work} for details): Their approach \DAVI (Deep Approximate Value Iteration) learned from scratch to solve arbitrary scrambled 3x3x3 cubes.

This work investigates whether TD-n-tuple learning with much lower computational demands can solve (or partially solve) Rubik's cube as well.

\subsection{Overview}
\label{sec:overview}

The General Board Game (GBG) learning and playing framework \citep{Konen2019b,Konen20b,Konen22a} was developed for education and research in AI. GBG allows applying the new algorithm easily to a variety of games. GBG is open source and available on  GitHub\footnote{\url{https://github.com/WolfgangKonen/GBG}}.
The main contribution of this paper is to take the TD-n-tuple approach from GBG \citep{Scheier2022} that was also successful on other games (Othello, ConnectFour) and to investigate this algorithm on various cube puzzles. We will show that it can solve the 2x2x2 cube perfectly and the 3x3x3 cube partly. At the same time it has drastically reduced computational requirements compared to \cite{mcaleer2019solving}. We will show that wrapping the base agent with an \textbf{MCTS wrapper}, as it was done by \cite{mcaleer2019solving} and \cite{Scheier2022}, is essential to reach this success.

This work is at the same time an in-depth tutorial how to represent a cube and its transformations within a computer program such that all types of cube operations can be computed efficiently. As another important contribution we will show how \textbf{symmetries} (Sec.~\ref{sec:symmetr}, \ref{sec:numSymmetry} and \ref{sec:resSymmetry}) applied to cube puzzles can greatly increase sample efficiency and performance.\\[0.2cm]

The rest of this paper is organized as follows: Sec.~\ref{sec:foundation} lays the foundation for Rubik's cube, its state representation, its transformations and its symmetries. In Sec.~\ref{sec:ntuples} we introduce n-tuple systems and how they can be used to derive policies for game-playing agents. Sec.~\ref{sec:represent-ntuple} defines and discusses several n-tuple representations for the cube. Sec.~\ref{sec:learning} presents algorithms for learning the cube: first the \DAVI algorithm of \cite{mcaleer2019solving,agostinelli2019solving} and then our n-tuple-based TD learning (with extensions TCL and MCTS).
In Sec.~\ref{sec:results} we present the results when applying our n-tuple-based TD learning method to the 2x2x2 and the 3x3x3 cube. Sec.~\ref{sec:rel-work} discusses related work and Sec.~\ref{sec:summary} concludes.

\newpage
\section{Foundations}
\label{sec:foundation}

\subsection{Conventions and Symbols}

We consider in this paper two well-known cube types, namely the 2x2x2 cube (pocket cube) and the 3x3x3 cube (Rubik's cube).

\label{sec:convent}
\subsubsection{Color arrangement}
Each cube consists of smaller \hypertarget{hrefCubie}{\textbf{cubies}}: 8 corner cubies for the 2x2x2 cube and 8 corner, 12 edge and 6 center cubies for the 3x3x3 cube. A corner cubie has 3 \hypertarget{hrefSticker}{\textbf{stickers}} of different color on its 3 faces. An edge cubie has two, a center cubie has one sticker.

We enumerate the 6 cube faces with \\
\hspace*{1.8cm}	(ULF) = (\textbf{U}p, \textbf{L}eft, \textbf{F}ront) and \\
\hspace*{1.8cm}	\hypertarget{hrefDRBcubie}{(DRB)} = (\textbf{D}own, \textbf{R}ight, \textbf{B}ack).

We number the 6 colors with 0,1,2,3,4,5. My cube has these six colors \\
\hspace*{1.8cm}	012 = wbo = (white,blue,orange) in the (ULF)-cubie\footnote{We run through the faces of a cubie in counter-clockwise orientation.}   and \\
\hspace*{1.8cm}	345 = ygr   = (yellow,green,red) in the opposing \DRB-cubie.

\begin{figure}[h]
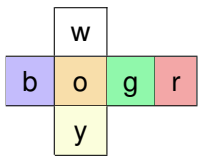

\renewcommand{\arraystretch}{1.35}
\centerline{
\begin{tabular}{|x|c|y|z|} \cline{2-2}
\multicolumn{1}{c|}{ } & w      &\multicolumn{2}{c}{ } \\ \hline
                     b & o\co & g & r \\  \hline
\multicolumn{1}{c|}{ } & y\ye &\multicolumn{2}{c}{ } \\ \cline{2-2}
\end{tabular}
}
\renewcommand{\arraystretch}{1.0}
\caption{The face colors of the default cube in flattened representation}
\label{fig:col-flattened}
\end{figure}

The solved cube in default position has colors (012345) for the faces (ULFDRB), i.e. the white color is at the \textbf{U}p face, blue at \textbf{L}eft, orange as \textbf{F}ront and so on. We can cut the cube such that up- and bottom-face can be folded away and have a flattened representation as shown in Figure~\ref{fig:col-flattened}.

\subsubsection{Twist and Rotation Symbols}
Twists of cube faces are denoted by uppercase letters U, L, F, D, R, B. Each of these twists means a $90^\circ$ counterclockwise rotation.\footnote{The rotation is counterclockwise when looking at the respective face} If U = U${}^{1}$ is a $90^\circ$ rotation, then U${}^{2}$ is a $180^\circ$ rotation and U${}^{3}$=U${}^{-1}$ is a $270^\circ$ rotation.

Whole-cube rotations are denoted by lowercase letters $u, \ell, f$. (We do not need $d, r, b$ here, because $d = u{}^{-1}, r = \ell^{-1}$ and so on.)

Further symbols like $f_c[i], \sL[i]$ that characterize a cube state  
will be explained in Sec.~\ref{sec:class_cs}.

\subsubsection{Twist Types}
Cube puzzles can have different twist types or twist metrics:
\begin{itemize}
	\item \textbf{\hypertarget{hrefQTM}{QTM} (quarter turn metric)}: only quarter twists are allowed: e.g. U${}^{1}$ and U${}^{-1}$.
	\item \textbf{\hypertarget{hrefHTM}{HTM} (half turn metric)}: quarter and half turns (twists) are allowed: e.g. U${}^{1}$, U${}^{2}$, U${}^{3}$.
\end{itemize}

By \textit{allowed} we mean what counts as one move. In QTM we can  realize U${}^{2}$ via U\,U as well, but it costs us 2 moves. In HTM, U${}^{2}$ counts as one move.

The twist type influences \GodsNumber and the branching factor of the game, see Sec.~\ref{sec:facts}.

\subsection{Facts about Cubes}
\label{sec:facts}
\subsubsection{2x2x2 Cube}
\label{sec:facts2x2}
	The \textbf{number of distinct states} for the 2x2x2 pocket cube is \citep{wikiPocketCube}
	\begin{equation}
		\frac{8! \cdot 3^7}{24} = 7!\cdot 3^6 = 3,674,160 \approx 3.6\cdot 10^6	
	\label{eq:numStates2x2}
	\end{equation}
Why this formula? –- We have 8 cubies which we can place in 8! ways on the 8 cube positions. Each but the last cubie has the freedom to appear in 3 orientations, which gives the factor $3^7$ (the last cubie is then in a fixed orientation, the other two orientations would yield illegal cube states). -- Each of these raw states has the (ygr)-cubie in any of the 24 possible positions. Or, otherwise speaking, each truly different state appears in 24 whole-cube rotations. To factor out the whole-cube rotations, we count only the states with (ygr)-cubie in its default position \DRB and divide the number of raw states by 24, q.e.d.
	
	\textbf{\hypertarget{hrefGodsNum}{God’s number}}: What is the minimal number of moves needed to solve any cube position? -- For the 2x2x2 pocket cube, it is 11 in \HTM (half-turn metric) and 14 in \QTM.
	
	\textbf{Branching factor}: $3\cdot 3 = 9$ in \HTM and $3\cdot 2 = 6$ in \QTM.

\subsubsection{3x3x3 Cube}
\label{sec:facts3x3}
	The \textbf{number of distinct states} for the 3x3x3 Cube is \citep{wikiRubiksCube}
	\begin{equation}
		\frac{8!\cdot 3^7\cdot 12!\cdot 2^{11}}{2} = 43,252,003,274,489,856,000 \approx 4.3\cdot 10^{19}	
	\label{eq:numStates3x3}
	\end{equation}
Why this formula? -- We have 8 corner cubies which we can place in 8! ways on the 8 cube positions. Each but the last cubie has the freedom to appear in 3 orientations, which gives the factor $3^7$. We have 12 edge cubies which we can place in 12! ways on the edge positions. Each but the last cubie has the freedom to appear in 2 orientations, which gives the factor $2^{11}$. 
The division by 2 stems from the fact, that neither alone two corner cubies may be swapped nor alone two edge cubies may be swapped. Instead, the number of such swaps must be even (factor 2). 

	\textbf{God’s Number}: What is the minimal number of moves needed to solve any cube position? –- For the 3x3x3 Rubik’s Cube, it is 20 in \HTM  (half-turn metric) and 26 in \QTM. This is a result from \cite{rokicki2014diameter}, see also \href{http://www.cube20.org/qtm/}{\url{http://www.cube20.org/qtm/}}.

	\textbf{Branching factor}: $6\cdot 3 = 18$ in \HTM and $6\cdot 2 = 12$ in \QTM.

\begin{figure}%
\centerline{
\renewcommand{\arraystretch}{1.75}
\begin{scriptsize}
\begin{tabular}{|x|x|c|c|y|y|z|z|} \cline{3-4}
\multicolumn{2}{c|}{ } & 3   & 2   &\multicolumn{4}{c}{ } \\ \cline{3-4}
\multicolumn{2}{c|}{ } & 0   & 1   &\multicolumn{4}{c}{ } \\ \hline
                 5 & 4 & 8\co&11\co& 18& 17& 23& 22\\  \hline
                 6 & 7 & 9\co&10\co& 19& 16& 20& 21\\  \hline
\multicolumn{2}{c|}{ } &14\ye&13\ye&\multicolumn{4}{c}{ } \\ \cline{3-4}
\multicolumn{2}{c|}{ } &15\ye&12\ye&\multicolumn{4}{c}{ } \\  \cline{3-4}
\end{tabular}
\end{scriptsize}
\renewcommand{\arraystretch}{1.0}
}
\caption{Sticker numbering for the 2x2x2 cube}
\label{fig:2x2x2stickers}
\end{figure}

\subsection{The Cube State}
\label{sec:class_cs}

A cube should be represented by objects in GBG in such a way that 
\begin{enumerate}[(a)]
	\item cube states that are equivalent are represented by identical objects
	\item if two cube states are equivalent, it should be easy to check this by comparing their objects
	\item cube transformations are easy to carry out on these objects.
\end{enumerate}

Condition (a) means that if two twist sequences lead to the same cube state (e.g. U$^{-1}$ and UUU), this should result also in identical objects. Condition (b) means, that the equality should be easy to check, given the objects. That is, a cube should \textit{not} be represented by its twist sequence.

A cube state is in GBG represented by abstract class \texttt{CubeState} and has two describing members
\begin{eqnarray}
		 f_c[i] &=& \fcol[i] \label{eq:fcol} \\
		 \sL[i] &=& \sloc[i] \label{eq:sloc}
\end{eqnarray}

$f_c[i] = \fcol[i]$ denotes the \textbf{f}ace \textbf{col}or at sticker location $i$. The color is one out of {0,1,2,3,4,5} for the colors {w,b,o,y,g,r}. 

$\sL[i] = \sloc[i]$ contains the \textbf{s}ticker \textbf{loc}ation of the \textit{\Sticker which is in position $i$ for the solved cube $d$}.

Members $f_c$ and $\sL$ are vectors with 24 (2x2x2 cube) or 48 (3x3x3 cube) elements where $i$ denotes the $i$th \Sticker location. 

The stickers are numbered in a certain way which is detailed in Figures~\ref{fig:2x2x2stickers} and 
\ref{fig:3x3x3stickers} for the flattened representations of the 2x2x2 and 3x3x3 cube, resp.

\begin{figure}%
\centerline{
\renewcommand{\arraystretch}{1.75}
\begin{scriptsize}
\begin{tabular}{|x|x|x|c|c|c|y|y|y|z|z|z|} \cline{4-6}
\multicolumn{3}{c|}{ } &  6   &  5   &  4   & \multicolumn{4}{c}{ } \\  \cline{4-6}
\multicolumn{3}{c|}{ } &  7   &      &  3   & \multicolumn{4}{c}{ } \\  \cline{4-6}
\multicolumn{3}{c|}{ } &  0   &  1   &  2   & \multicolumn{4}{c}{ } \\  \hline
             10& 9 & 8 & 16\co& 23\co& 22\co& 36& 35& 34& 46& 45& 44\\  \hline
             11&   & 15& 17\co&   \co& 21\co& 37&   & 33& 47&   & 43\\  \hline
             12& 13& 14& 18\co& 19\co& 20\co& 38& 39& 32& 40& 41& 42\\  \hline
\multicolumn{3}{c|}{ } & 28\ye& 27\ye& 26\ye& \multicolumn{4}{c}{ } \\  \cline{4-6}
\multicolumn{3}{c|}{ } & 29\ye&   \ye& 25\ye& \multicolumn{4}{c}{ } \\  \cline{4-6}
\multicolumn{3}{c|}{ } & 30\ye& 31\ye& 24\ye& \multicolumn{4}{c}{ } \\  \cline{4-6}
\end{tabular}
\end{scriptsize}
\renewcommand{\arraystretch}{1.0}
}
\caption{Sticker numbering for the 3x3x3 cube. We do not number the center \Cubies, they stay invariant under twists.}
\label{fig:3x3x3stickers}
\end{figure}

In principle, one of the two members $f_c$ and $\sL$ would be sufficient to characterize a state, since the \textbf{fcol-sloc-relation} 
\begin{equation}
		f_c[\sL[i]] = d.f_c[i]
\label{eq:fcolsloc}
\end{equation}
holds, where $d$ denotes the default cube. 
This is because $\sL[i]$ transports the sticker $i$ of the default cube $d$ to location $\sL[i]$, i.e. it has the color $d.f_c[i]$. 
That is, we can easily calculate $f_c$ given $\sL$. With some more effort, it is also possible to calculate $\sL$ given $f_c$ (see Appendix~\ref{app:calc_s_from_f}). Although one of these members $f_c$ and $\sL$ would be sufficient, we keep both because this allows to better perform assertions or cross checks during transformations.

Sometime we need the inverse function $\sL^{-1}[i]$: \textit{Which sticker is at location $i$?} It is easy to calculate $\sL^{-1}$ given $\sL$ with the help of the relation:
\begin{equation}
		\sL^{-1}[\sL[i]] = i
\label{eq:sloc-inv}
\end{equation}
(Note that it is \textit{not} possible to invert $f_c$, because the face coloring function is not bijective.)

\subsection{Transformations}
\label{sec:transform}

\begin{table}%
\caption{The three relevant twists for the 2x2x2 cube}
\label{tab:twist}
\centerline{
\begin{scriptsize}
\tabcolsep=0.08cm			
\renewcommand{\arraystretch}{1.1}
\begin{tabular}{|l|c||a|a|a|a||c|c|c|c||a|a|a|a||c|c|c|c||a|a|a|a||c|c|c|c||} \hline
				&			    &0 &1	&2  &3	 &4	&5	&6 &7	 &8	&9	&10	&11	 &12 &13 &14 &15	 &16 &17	&18	&19	 &20	&21	&22	&23 \\ \hline\hline
U twist &	 $T$  	&1 &2	&3	&0	 &11&8	&6 &7  &18&9	&10	&17  &12 &13 &14 &15   &16 &22	&23	&19  &20	&21	&4	&5  \\ \hline
L twist	&	 $T$	  &22&1	&2 &21   &5	&6	&7 &4	 &3	&0	&10	&11  &12 &13 &8	 &9	   &16 &17	&18	&19	 &20	&14	&15	&23 \\ \hline
F twist	&	 $T$  	&7 &4	&2 &3	   &14&5	&6 &13 &9	&10	&11	&8	 &12 &18 &19 &15   &16 &17	&0	&1	 &20	&21	&22	&23 \\ \hline\hline
U$^{-1}$&	$T^{-1}$&3 &0	&1 &2	   &22&23	&6 &7	 &5	&9	&10	&4	 &12 &13 &14 &15   &16 &11	&8	&19	 &20	&21	&17	&18 \\ \hline
L$^{-1}$&	$T^{-1}$&	9&1	&2 &8	   &7	&4	&5 &6	 &14&15	&10	&11  &12 &13 &21 &22   &16 &17	&18	&19	 &20	&3	&0	&23 \\ \hline
F$^{-1}$&	$T^{-1}$&18&19&2 &3	   &1	&5	&6 &0	 &11&8	&9	&10  &12 &7	 &4	 &15   &16 &17	&13	&14	 &20	&21	&22	&23 \\ \hline
\end{tabular}
\renewcommand{\arraystretch}{1.0}
\end{scriptsize}
}
\end{table}

\subsubsection{Twist Transformations}
\label{sec:twist}
Each basic twist is a counterclockwise\footnote{The rotation is counterclockwise when looking at this face.} rotation of a face by $90^\circ$.  
Table~\ref{tab:twist} shows the 2x2x2 transformation functions for three basic twists. Each twist transformation can be coded in two forms:
\begin{enumerate}
	\item $T[i]$ (forward transformation): Which is the new location for the \Sticker being at $i$ before the twist?
	\item $T^{-1}[i]$ (inverse transformation): Which is the (parent) location of the \Sticker that lands in $i$ after the twist? 
\end{enumerate}

Example (read off from column $0$ of Table~\ref{tab:twist}): The L-twist transports sticker at $0$ to $22$: $T[0]=22$.  The (parent) sticker being at location $9$ before the L-twist comes to location $0$ after the twist: $T^{-1}[0]=9$. Likewise, for the U-twist we have $T[0]=1$ and $T^{-1}[0]=3$. We show in Fig.~\ref{fig:utwist} the default cube after twist U${}^{1}$.

How can we apply a twist transformation to a cube state programmatically? -- We denote with $f_c'$ and $\sL'$ the new states for $f_c$ and $\sL$ after transformation. The following relations allow to calculate the transformed cube state:
\begin{eqnarray}
		 f_c'[i]         						 &=& f_c[T^{-1}[i]] \label{eq:utwist_f}\\
		 \sL'[\sL^{-1}[i]] &=& T[i]         	\label{eq:utwist_s}
\end{eqnarray}

Eq.~\eqref{eq:utwist_f} says: The new color for sticker $0$ is the color of the sticker which moves into location $0$ ($f_c[9]$ in the case of an L-twist). To explain Eq.~\eqref{eq:utwist_s}, we first note that $\sL^{-1}[i]$ is the sticker being at $i$ before the transformation. Then, Eq.~\eqref{eq:utwist_s} says: \glqq The new location for the sticker being at $i$ before the transformation is $T[i]$.\grqq\ For example, the L-twist transports the current sticker at location $0$ to the new location $T[0]=22$, i.\,e. $\sL'[0]=22$.

\begin{figure}%
\centerline{
\renewcommand{\arraystretch}{1.75}
\begin{scriptsize}
\begin{tabular}{|c|c|c|c|c|c|c|c|} \cline{3-4}
\multicolumn{2}{c|}{ } & 2   & 1   &\multicolumn{4}{c}{ } \\ \cline{3-4}
\multicolumn{2}{c|}{ } & 3   & 0   &\multicolumn{4}{c}{ } \\ \hline
            23\re&22\re& 5\bl& 4\bl&  8\co& 11\co& 18\gr& 17\gr\\  \hline
             6\bl& 7\bl& 9\co&10\co& 19\gr& 16\gr& 20\re& 21\re\\  \hline
\multicolumn{2}{c|}{ } &14\ye&13\ye&\multicolumn{4}{c}{ } \\ \cline{3-4}
\multicolumn{2}{c|}{ } &15\ye&12\ye&\multicolumn{4}{c}{ } \\  \cline{3-4}
\end{tabular}
\end{scriptsize}
\renewcommand{\arraystretch}{1.0}
}
\caption{The default 2x2x2 cube after twist U$^{1}$}
\label{fig:utwist}
\end{figure}

For the 2x2x2 cube, these 3 twists U, L, F are sufficient, because D=U$^{-1}$, R=L$^{-1}$, B=F$^{-1}$. This is because the 2x2x2 cube has no center \Cubies. For the 3x3x3 cube, we need all 6 twists U, L, F, D, R, B because this cube has center cubies.

\begin{table}[bp]%
\caption{The U twist for the 3x3x3 cube}
\label{tab:Utwist3x3}
\centerline{
\begin{scriptsize}
\tabcolsep=0.08cm			
\renewcommand{\arraystretch}{1.1}
\begin{tabular}{|l|c||a|a|a|a||c|c|c|c||a|a|a|a||c|c|c|c||a|a|a|a||c|c|c|c||} \hline
				&			    &0 &1	&2  &3	 &4	&5	&6 &7	 &8	&9	&10	&11	 &12 &13 &14 &15	 &16 &17	&18	&19	 &20	&21	&22	&23 \\ \hline\hline
U twist &	 $T$  	&2 &3	&4	&5	 &6 &7  &0 &1  &22&23 &16	&11  &12 &13 &14 &15   &36 &17	&18	&19  &20	&21	&34 &35 \\ \hline
				&			    &24&25&26 &27  &28&29 &30&31 &32&33 &34	&35	 &36 &37 &38 &39	 &40 &41	&42	&43	 &44	&45	&46	&47 \\ \hline\hline
U twist	&	 $T$	  &24&25&26 &27  &28&29 &30&31 &32&33 &44	&45  &46 &37 &38 &39   &40 &41	&42	&43	 &8 	&9 	&10	&47 \\ \hline

\end{tabular}
\renewcommand{\arraystretch}{1.0}
\end{scriptsize}
}
\end{table}

In any case, we will show in Sec.~\ref{sec:wcr} that only one row in Table~\ref{tab:twist} or Table~\ref{tab:Utwist3x3}, say $T$ for the U-twist, has to be known or established 'by hand'. All other twists and their inverses can be calculated programmatically with the help of Eqs.~\eqref{eq:T-inv}-\eqref{eq:twistsFromWCR_B} that will be derived in Sec.~\ref{sec:wcr}.

\hypertarget{hrefNormalize2x2}{\paragraph{\hspace*{1cm}\textit{Normalizing the 2x2x2 Cube}}}
As stated above, the 3 twists U, L, F are sufficient for the 2x2x2 cube. Therefore, the \DRB-cubie will never leave its place, whatever the twist sequence formed by U, L, F is. The (DRB)-cubie has the stickers (12, 16, 20), and we can check in Table~\ref{tab:twist} that columns (12, 16, 20) are always invariant. If we have an arbitrary initial 2x2x2 cube state, we can normalize it by applying a \WholeCubeRotation such that the (ygr)-cubie moves to the (DRB)-location.

\hypertarget{hrefNormalize3x3}{\paragraph{\hspace*{1cm}\textit{Normalizing the 3x3x3 Cube}}}
In the case of the 3x3x3 cube, all center cubies will be not affected by any twist sequence. Therefore, we normalize a 3x3x3 cube state by applying initially a \WholeCubeRotation such that the center cubies are in their normal position (i.e. white up, blue left and so on).

\hypertarget{hrefWCR}{\subsubsection{Whole-Cube Rotations (WCR)}}
\label{sec:wcr}
Each basic \textbf{whole-cube rotation} (WCR) is a counterclockwise rotation of the whole cube around the $u,l,f$-axis by $90^\circ$.  
Table~\ref{tab:basic-wcr} shows two of the 2x2x2 transformation functions for basic whole-cube rotations. Each rotation can be coded in two forms:

\begin{enumerate}
	\item $T[i]$ (forward transformation): Which is the new location for the \Sticker being at $i$ before the twist?
	\item $T^{-1}[i]$ (inverse transformation): Which is the (parent) location of the \Sticker that lands in $i$ after the twist? 
\end{enumerate}

\begin{table}%
\caption{Two basic whole-cube rotations for the 2x2x2 cube}
\label{tab:basic-wcr}
\centerline{
\begin{scriptsize}
\tabcolsep=0.08cm			
\renewcommand{\arraystretch}{1.1}
\begin{tabular}{|c|c||j|j|j|j||c|c|c|c||j|j|j|j||c|c|c|c||j|j|j|j||c|c|c|c||} \hline
				&			       &0 &1	&2  &3	 &4	&5	&6 &7	 &8	&9	&10	&11	 &12 &13 &14 &15	 &16 &17	&18	&19	 &20	&21	&22	&23 \\ \hline\hline
$u$ rotation &	 $T$ &1 &2	&3	&0	 &11&8	&9 &10 &18&19&16	&17  &15 &12 &13 &14   &21 &22	&23	&20  & 6	& 7	&4	&5  \\ \hline
$f$ rotation &	 $T$ &7 &4	&5 &6	   &14&15&12&13 &9	&10	&11	&8	 &17 &18 &19 &16   & 2 & 3	&0	&1	 &23	&20	&21	&22 \\ \hline\hline
$u^{-1}$&	$T^{-1}$	 &3 &0	&1 &2	   &22&23	&20&21 &5	&6	& 7	&4	 &13 &14 &15 &12   &10 &11	&8	&19	 &19	&16	&17	&18 \\ \hline
$f^{-1}$&	$T^{-1}$	 &18&19&16&17   &1	&2	&3 &0	 &11&8	&9	&10  & 6 &7	 &4	 & 5   &15 &12	&13	&14	 &21	&22	&23	&20 \\ \hline
\end{tabular}
\renewcommand{\arraystretch}{1.0}
\end{scriptsize}
}
\end{table}

Besides the basic rotation $u$ there is also $u^2$ ($180^\circ$) and $u^3=u^{-1}$ ($270^\circ = - 90^\circ$).

All whole-cube rotations can be generated from these two forward rotations $u$ and $f$: First, we calculate the inverse transformations via
\begin{equation}
		T^{-1}[T[i]] = i
\label{eq:T-inv}
\end{equation}
where $T$ is a placeholder for $u$ or $f$. Next, we calculate the missing base rotation $\ell$ (counter-clockwise around the left face) as
\begin{equation}
		\ell = fuf^{-1}
\label{eq:wcr-left}
\end{equation}
We use here the programm-code-oriented notation \hypertarget{hrefFTF}{\textbf{\glqq first trafo first\grqq}}: Eq.~\eqref{eq:wcr-left} reads as \glqq first $f$, then $u$, then $f^{-1}$\grqq.\footnote{In programm code the relation would read \texttt{cs.fTr(1).uTr().fTr(3)}. This is \textbf{\glqq first trafo first\grqq}, because each transformation is applied to the cube state object to the left and returns the transformed cube state object.}

The other basic whole-cube rotations $d,r,b$ are not needed, because $d=u^{-1}, r=\ell^{-1}$ and $b=f^{-1}$.

The basic whole-cube rotations are rotations of the whole cube around just one axis. But there are also composite whole-cube rotations which consists of a sequence of basic rotations.

How many different (composite) rotations are there for the cube? -- A little thought reveals that there are 24 of them: To be specific, we consider the default cube where we have 4 rotations with the white face up, 4 with the blue face up, and so on. In total we have $6\cdot 4=24$ rotations since there are 6 faces.
Table~\ref{tab:wcr} lists all of them, togehter with the \WCR numbering convention used in GBG.

\begin{table}%
\caption{All 24 whole-cube rotations (in \firstTrafoFirst notation)}
\label{tab:wcr}
\centerline{
\begin{tabular}{lccccc} \hline
number & first rotation & $\ast\, u^{0}$& $\ast\, u^{1}$& $\ast\, u^{2}$& $\ast\, u^{3}$ \\ \hline\hline
00-03	 & \textit{id} (white up)	&\textit{id}	  &$u$	&$u^2$	&$u^3$  \\
04-07  &	$f$ (green up)	      &$f$						&$fu$	&$fu^2$	&$fu^3$ \\
08-11  &	$f^2$ (yellow up)			&$f^2$					&$f^2u$	&$f^2u^2$	&$f^2u^3$ \\
12-15  &	$f^{-1}$ (blue up)		&$f^{-1}$				&$f^{-1}u$	&$f^{-1}u^2$	&$f^{-1}u^3$ \\
16-19  &	$\ell$ (orange up)		&$\ell$					&$\ell u$	&$\ell u^2$	&$\ell u^3$ \\
20-23  &	$\ell^{-1}$ (red up)	&$\ell^{-1}$		&$\ell^{-1}u$	&$\ell^{-1}u^2$	&$\ell^{-1}u^3$ \\ \hline\hline
\end{tabular}
}
\end{table}

Sometimes we need the inverse whole-cube rotations which are given in Table~\ref{tab:wcr-inverse}. In this table, we read for example from the element with number 5, that the \WCR with key 5 (which is $fu$ according to Table~\ref{tab:wcr}) has the inverse WCR $\ell u^3$ such that
$$
		fu \, \ell u^3 = \textit{id}
$$
holds.

For convenience, we list in Table~\ref{tab:invkey} the <Key, InverseKey> relation. For example, the trafo with Key=5 ($fu$) has the inverse trafo with InverseKey=19 ($\ell u^3$). Note that there are 10 whole-cube rotations which are their own inverse.

\begin{table}%
\caption{The 24 \textit{inverse} whole-cube rotations (in \firstTrafoFirst notation)}
\label{tab:wcr-inverse}
\centerline{
\begin{tabular}{lccccc} \hline
number & first rotation & $\ast\, u^{0}$& $\ast\, u^{1}$& $\ast\, u^{2}$& $\ast\, u^{3}$ \\ \hline\hline
00-03	 & \textit{id} (white up)	&\textit{id}	&$u^3$	&$u^2$	&$u^1$  \\
04-07  &	$f$ (green up)	      &$f^{-1}$			&$\ell u^3$	&$fu^2$	&$\ell^{-1}u$ \\
08-11  &	$f^2$ (yellow up)			&$f^2$				&$f^2u$	&$f^2u^2$	&$f^2u^3$ \\
12-15  &	$f^{-1}$ (blue up)		&$f$					&$\ell^{-1}u^3$	&$f^{-1}u^2$	&$\ell u$ \\
16-19  &	$\ell$ (orange up)		&$\ell^{-1}$	&$f^{-1}u^3$	&$\ell u^2$	&$fu$ \\
20-23  &	$\ell^{-1}$ (red up)	&$\ell$				&$fu^{-1}$	&$\ell^{-1}u^2$	&$f^{-1}u$ \\ \hline\hline
\end{tabular}
}
\end{table}

\begin{table}%
\caption{Whole-cube rotations: <Key, InverseKey> relation}
\label{tab:invkey}
\centerline{
\begin{scriptsize}
\tabcolsep=0.08cm			
\renewcommand{\arraystretch}{1.1}
\begin{tabular}{|c||j|j|j|j||c|c|c|c||j|j|j|j||c|c|c|c||j|j|j|j||c|c|c|c||} \hline
key 		&	0 &1	&2  &3	 &4	&5	&6 &7	  &8 &9	&10	&11	   &12 &13 &14 &15	 &16 &17	&18	&19  	 &20	&21	&22	&23 \\ \hline\hline
inv key &	0 &3	&2	&1	 &12&19&6 &21   &8 &9 &10	&11    & 4 &23 &14 &17   &20 &15	&18	&05    &16	& 7	&22 &13 \\ \hline
\end{tabular}
\renewcommand{\arraystretch}{1.0}
\end{scriptsize}
}
\end{table}

\paragraph{Generating all twists from U twist}
With the help of \WCRs we can generate the other twists from the U twist only: We simply rotate the face that we want to twist to the up-face, apply the U twist and rotate back. This reads in \firstTrafoFirst notation:
\begin{eqnarray}
		L = f^{-1} U f \label{eq:twistsFromWCR_L}\\
		F = \ell U \ell^{-1} \\
		D = f^2 U f^2\\
		R = f U f^{-1} \\
		B = \ell^{-1} U \ell  \label{eq:twistsFromWCR_B}
\end{eqnarray}

Thus, given the U twist from Table~\ref{tab:twist} or Table~\ref{tab:Utwist3x3} and the basic \WCRs given in Table~\ref{tab:basic-wcr} and Eq.~\eqref{eq:wcr-left}, we can calculate all other forward transformations with the help of Eqs.~\eqref{eq:twistsFromWCR_L}--\eqref{eq:twistsFromWCR_B}. Then, all inverse transformations are calculable with the help of Eq.~\eqref{eq:T-inv}.

\subsubsection{Color Transformations}
\label{sec:colortrans}

Color transformations are special transformations that allow to discover non-trivial symmetric (equivalent) states. 

One way to describe a color transformation is to select a valid color permutation and to paint each sticker with the new color according to this color permutation. This is of course nothing one can do with a real cube without destroying or altering it, but it is a theoretical concept leading to an equivalent state.

Another way of looking at it is to record the twist sequence that leads from the default cube to a certain scrambled cube state. Then we go back to the default cube, make at first a whole-cube rotation (leading to a color-transformed default cube) and then apply the recorded twist sequence to the color-transformed default cube.

In any case, the transformed cube will be usually not in its normal position, so we apply finally a \href{\#hrefNormalize2x2}{normalizing operation} to it. 

What are valid color permutations? -- These are permutations of the cube colors reachable when applying one of the available 24 \WCRs (Table~\ref{tab:wcr}) to the default cube.
For example, if we apply WCR $f$ (number 04) to the default cube, we get
\vspace{0.1cm}

\begin{figure}[h]
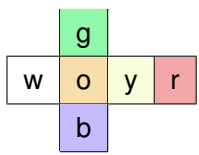
%
\renewcommand{\arraystretch}{1.30}
\centerline{
\begin{tabular}{|c|c|v|z|} \cline{2-2}
\multicolumn{1}{c|}{ } & g\gr &\multicolumn{2}{c}{ } \\ \hline
                     w & o\co & y & r \\  \hline
\multicolumn{1}{c|}{ } & b\bl &\multicolumn{2}{c}{ } \\ \cline{2-2}
\end{tabular}
}
\renewcommand{\arraystretch}{1.0}
\vspace{0.1cm}
\caption{The color transformation according to \WCR $f$ (number 04)}
\label{fig:colorTrafo}
\end{figure}

\noindent
that is, g (green) is the new color for each up-sticker that was w (white) before and so on. The colors o and r remain untouched under this color permutation. 
[However, other transformations like $fu$, $fu^2$ and $fu^3$ will change every color.]

How can we apply a color transformation to a cube state programmatically? -- We denote with $f'$ and $\sL'$ the new states for $f$ and $\sL$ after transformation. The following relations allow to calculate the transformed cube state:
\begin{eqnarray}
		 f_c'[i]       							 &=& c[f_c[i]] \label{eq:coltraf_f}\\
		 \sL'[\sL^{-1}[i]] &=& T[i]      \label{eq:coltraf_s}
\end{eqnarray}
where $c[]$ is the 6-element color trafo vector (holding the new colors for current colors 0:w, 1:b, ..., 5:r) and $T$ is the 24- or 48-element vector of the \WCR that produces this color transformation. Eq.~\eqref{eq:coltraf_f} is simple: If a certain sticker has color 0 (w, white) before the color transformation, then it will get the new color $c[0]$, e.g. 4 (g, green), after the transformation. Eq.~\eqref{eq:coltraf_s} looks complicated, but it has a similar meaning as in the twist trafo: Take $i=0$ as example: The new place for the sticker being at 0 before the trafo (and coming from $\sL^{-1}[0]$) is $T[0]$. Therefore, we write the number $T[0]$ into $\sL'[\sL^{-1}[0]]$.

\begin{figure}%
\centerline{
\renewcommand{\arraystretch}{1.75}
\begin{scriptsize}
\begin{tabular}{|c|c|c|c|c|c|c|c|} \cline{3-4}
\multicolumn{2}{c|}{ } & 2   & 1   &\multicolumn{4}{c}{ } \\ \cline{3-4}
\multicolumn{2}{c|}{ } & 3   & 0   &\multicolumn{4}{c}{ } \\ \hline
            23\re&22\re& 5\bl& 4\bl&  8\co& 11\co& 18\gr& 17\gr\\  \hline
             6\bl& 7\bl& 9\co&10\co& 19\gr& 16\gr& 20\re& 21\re\\  \hline
\multicolumn{2}{c|}{ } &14\ye&13\ye&\multicolumn{4}{c}{ } \\ \cline{3-4}
\multicolumn{2}{c|}{ } &15\ye&12\ye&\multicolumn{4}{c}{ } \\  \cline{3-4}
\end{tabular}
\end{scriptsize}
\renewcommand{\arraystretch}{1.0}
}
\caption{The cube of Fig.~\ref{fig:utwist} before color transformation.}
\label{fig:state-beforeCT}
\end{figure}

\begin{figure}%
\centerline{
\renewcommand{\arraystretch}{1.75}
\begin{tabular}{cc}
\begin{minipage}[t]{8cm}
\hspace*{1.2cm}
\begin{scriptsize}
\begin{tabular}{|c|c|c|c|c|c|c|c|} \cline{3-4}
\multicolumn{2}{c|}{ } &16\gr&19\gr&\multicolumn{4}{c}{ } \\ \cline{3-4}
\multicolumn{2}{c|}{ } &17\gr&18\gr&\multicolumn{4}{c}{ } \\ \hline
            20\re&23\re& 2   & 1   & 11\co& 10\co& 13\ye& 12\ye\\  \hline
             3   & 0   & 8\co& 9\co& 14\ye& 15\ye& 21\re& 22\re\\  \hline
\multicolumn{2}{c|}{ } & 4\bl& 7\bl&\multicolumn{4}{c}{ } \\ \cline{3-4}
\multicolumn{2}{c|}{ } & 5\bl& 6\bl&\multicolumn{4}{c}{ } \\  \cline{3-4}
\end{tabular}
\end{scriptsize}
\end{minipage}
&
\begin{minipage}[t]{8cm}
\hspace*{1.2cm}
\begin{scriptsize}
\begin{tabular}{|c|c|c|c|c|c|c|c|} \cline{3-4}
\multicolumn{2}{c|}{ } & 8\co& 2   &\multicolumn{4}{c}{ } \\ \cline{3-4}
\multicolumn{2}{c|}{ } & 9\co& 1   &\multicolumn{4}{c}{ } \\ \hline
             4\bl& 7\bl&14\ye&11\co& 18\gr& 17\gr& 23\re&  0   \\  \hline
             5\bl& 6\bl&15\ye&10\co& 19\gr& 16\gr& 20\re&  3   \\  \hline
\multicolumn{2}{c|}{ } &21\re&13\ye&\multicolumn{4}{c}{ } \\ \cline{3-4}
\multicolumn{2}{c|}{ } &22\re&12\ye&\multicolumn{4}{c}{ } \\  \cline{3-4}
\end{tabular}
\end{scriptsize}
\end{minipage}
\\
(a) & (b) \\
\end{tabular}
\renewcommand{\arraystretch}{1.0}
}
\caption{The cube of Fig.~\ref{fig:state-beforeCT} with color transformation from Fig~\ref{fig:colorTrafo}: (a) before normalization, (b) after normalization.}
\label{fig:state-afterCT}
\end{figure}

A \textbf{color transformation example} is shown in Figs.~\ref{fig:state-beforeCT} and \ref{fig:state-afterCT}. Fig.~\ref{fig:state-beforeCT} is just a replication of Fig.~\ref{fig:utwist} showing a default cube after U${}^1$ twist. The color transformation number 04 applied to the cube of Fig.~\ref{fig:state-beforeCT} is shown in Fig.~\ref{fig:state-afterCT} (a)-(b) in two steps: 
\begin{enumerate}[(a)]
	\item The stickers are re-painted and re-numbered (white becomes green, blue becomes white and so on). The structure of coloring is the same as in Fig.~\ref{fig:state-beforeCT}. Now the \DRB-cubie is no longer the (ygr)-cubie, it does not carry the numbers (12,16,20). 
	\item We apply the proper \WCR that brings the (ygr)-cubie back to the \DRB-location. Compared to (a), each 4-sticker cube face is just rotated to another face, but not changed internally. We can check that the (DRB)-location now carries again the numbers (12,16,20), as in Fig.~\ref{fig:state-beforeCT} and as it should for a normalized cube.
\end{enumerate}

\subsection{Symmetries}
\label{sec:symmetr}

Symmetries are transformations of the game state (and the attached action, if applicable) that lead to equivalent states. That is, if $s$ is a certain state with value $V(s)$, then all  states $s_{sym}$ being symmetric to $s$ have the same value $V(s_{sym})=V(s)$ because they are equivalent. \textit{Equivalent} means: If $s$ can be solved by a twist sequence of length $n$, then $s_{sym}$ can be solved by an equivalent twist sequence of same length $n$.

In the case of Rubik's cube, all whole-cube rotations (\WCRs) are symmetries because they do not change the value of a state. But whole-cube rotations are 'trivial' symmetries because they are usually factored out by the normalization of the cube: After \href{\#hrefNormalize2x2}{2x2x2 cube normalization}, which brings the (ygr)-cubie in a certain position, or after \href{\#hrefNormalize3x3}{3x3x3 cube normalization}, which brings the center cubies in certain faces, all \WCR-symmetric states are transformed to the same state.

Non-trivial symmetries are all color transformations (Sec.~\ref{sec:colortrans}): In general, color transformations transform a state $s$ to a truly different state $s_{sym}$, even after \href{\#hrefNormalize2x2}{cube normalization}.\footnote{In rare cases -- e.g. for the solved cube -- the transformed state may be identical to $s$ or to another symmetry state, but this happens seldom for sufficiently scrambled cubes, see Sec.~\ref{sec:numSymmetry}.}
Since there are 24 color transformations in Rubik's cube, there are also 24 non-trivial symmetries (including self). 

Symmetries are useful to learn to solve Rubik's cube for two reasons: (a) to accelerate  learning  and (b) to smooth an otherwise noisy value function. 
\begin{enumerate}[(a)]
	\item \textbf{Accelerated learning}: If a state $s$ (or state-action pair) is observed, not only the weights activated by that state are updated, but also the weights of all symmetric states $s_{sym}$, because they have the same $V(s_{sym})=V(s)$ and thus the same reward. In this way, a single observed sample is connected with more weight updates (better sample efficiency).
	\item \textbf{Smoothed value function}:
By this we mean that the value function $V(s)$ is replaced by 
\begin{equation}
			V^{(sym)}(s) = \frac{1}{|\mathfrak{F}_s|} \sum_{s' \in \mathfrak{F}_s} V(s')
\label{eq:Vsym}
\end{equation}
where $\mathfrak{F}_s$ is the set of states being symmetric to $s$. If $V(s)$ were the ideal value function, both terms $V(s)$ and $V^{(sym)}(s)$ would be the same.\footnote{because all $V(s')$ in Eq.~\eqref{eq:Vsym} are the same for an ideal $V$} But in a real n-tuple network, $V(s)$ is non-ideal due to n-tuple-noise (cross-talk from other states that activate the same n-tuple LUT entries). If we average over the symmetric states $s' \in \mathfrak{F}_s$, the noise will be dampened.
\end{enumerate}

The downside of symmetries is their computational cost: In the case of Rubik's cube, the calculation of color transformations is a costly operation. On the other hand, the number of necessary training episodes to reach a certain performance may be reduced. 
In the end, the use of symmetries may pay off, because the total training time may be reduced as well. In any case, we will have a better sample efficiency, since we learn more from each observed state or state-action pair. Secondly, the smoothing effect introduced with Eq.~\eqref{eq:Vsym} can lead to better overall performance, because the smoothed value function provides a better guidance on the path towards the solved cube. 

In order to balance computation time, GBG offers the option to select with \texttt{nSym} the number of symmetries actually used.
If we specify for example \texttt{nSym}\,=\,8 in GBG's Rubik's cube implementation, then  the state itself and 8\,--\,1\,=\,7 random other (non-id) color transformations will be selected. The resulting set $\mathfrak{F}_s$ of 8 states is then used for weight update and value function computation.

\section{N-Tuple Systems}
\label{sec:ntuples}
N-tuple systems coupled with TD were first applied to game learning by \cite{Lucas08}, although n-tuples were already introduced by ~\cite{bledsoe1959pattern} for character recognition purposes. The remarkable success of n-tuples in learning to play Othello~\citep{Lucas08} motivated other authors to benefit from this approach for a number of other games. 

\begin{figure}%
\centerline{
	\includegraphics[width=0.5\columnwidth]{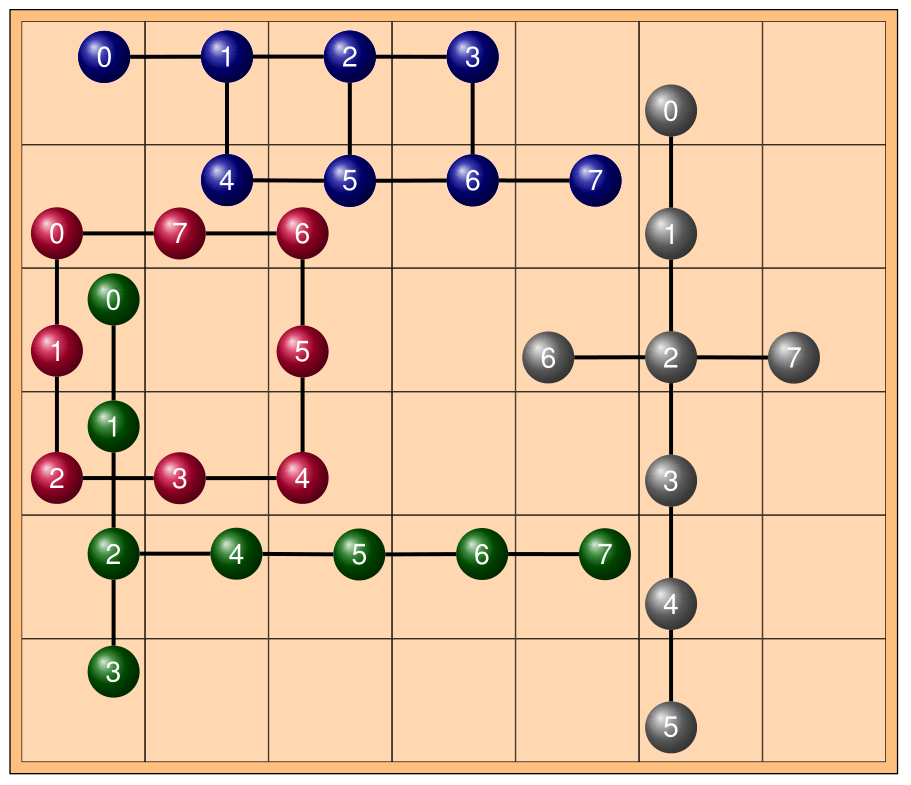}		
} 
\caption{Example n-tuples: We show 4 random-walk 8-tuples on a 6x7 board. The tuples are selected manually to show that not only snake-like shapes are possible, but also bifurcations or cross shapes. Tuples may or may not be symmetric.}
\label{fig:ntuple01}%
\end{figure}

The main goal of n-tuple systems is to map a highly non-linear function in a low dimensional space to a high dimensional space where it is easier to separate `good' and `bad' regions. This can be compared to the kernel trick of support-vector machines. An n-tuple is defined as a sequence of $n$ cells of the board. Each cell can have $m$ positional values representing the possible states of that cell.\footnote{A typical example is a 2-player board game, where we usually have 3 positional values \{0: empty, 1: player1, 2: player2 \}. But other, user-defined values are possible as well.} Therefore, every n-tuple will have a (possibly large) look-up table indexed in form of an $n$-digit number in base $m$. Each entry corresponds to a feature and carries a trainable weight. An n-tuple system is a system consisting of $k$ n-tuples. 
As an example we show in Fig.~\ref{fig:ntuple01} an n-tuple system consisting of four 8-tuples. 


Let $\bfTheta$ be the vector of all weights $\theta_i$ of the n-tuple system.\footnote{The index $i$ indexes three qualities: an n-tuple, a cell in this n-tuple and a positional value for this cell.} The length of this vector may be large number, e.g. $m^n k$, if all $k$ n-tuples have the same length $n$ and each cell has $m$ positional values. 
Let $\mathbf{\Phi}(s)$ be a binary vector of the same length representing the feature occurences in state $s$ (that is, $\mathbf{\Phi}_i(s)=1$ if in state $s$ the cell of a specific n-tuple as indexed by $i$ has the positional value as indexed by $i$, $\mathbf{\Phi}_i(s)=0$ else). The value function of the n-tuple network given state $s$ is 
\begin{equation}
		V(s) = \sigma \left( \mathbf{\Phi}(s)\cdot \bfTheta\right)
\label{eq:valueNtuple}
\end{equation}
with transfer function $\sigma$ which may be a sigmoidal function or simply the identity function.

An agent using this n-tuple system derives a policy from the value function in Eq.~\eqref{eq:valueNtuple} as follows: Given state $s$ and the set $A(s)$ of available actions in state $s$, it applies with a forward model $f$
every action $a \in A(s)$ to state $s$, yielding the next state $s' = f(s,a)$. Then it selects the action that maximizes $V(s')$.


Each time a new agent is constructed, all n-tuples are either created in fixed, user-defined positions and shapes, or they are formed by \textit{random walk}. In a \textit{random walk}, all cells are placed randomly with the constraint that each cell must be adjacent\footnote{The form of adjacency, e.~g. 4- or 8-point neighborhood or any other (might be cell-dependent) form of adjacency, is user-defined.} to at least one other cell in the n-tuple.

Agent training proceeds in the TD-n-tuple algorithm as follows:
Let $s'$ be the actual state generated by the agent and let $s$ be the previous state generated by this agent. TD(0) learning 
adapts the value function with model parameters $\bfTheta$ through \citep{SuttBart98}
\begin{equation}
		\bfTheta \leftarrow \bfTheta + \alpha\delta\mathbf{\nabla_{\bfTheta}} V(s)
\label{eq:theta}
\end{equation}
Here, $\alpha$ is the learning rate and $V$ is in our case the n-tuple value function of  Eq.~\eqref{eq:valueNtuple}. $\delta$ is the usual TD error \citep{SuttBart98} after the agent has acted and generated $s'$:
\begin{equation}
		\delta = r+\gamma V(s') - V(s)
\label{eq:TDdelta}
\end{equation}
where the sum of the first two terms, reward $r$ plus the discounted value $\gamma V(s')$, is the desirable target for $V(s)$.

\section{N-Tuple Representions for the Cube}
\label{sec:represent-ntuple}

In order to apply n-tuples to cubes, we have to define a board in one way or the other on which we can place the n-tuples. This is not as straightforward as in other board games, but we are free to invent abstract boards. Once we have defined a board, we can number the board cells $k=0,\ldots,K-1$ and translate a cube state into a BoardVector: A \hypertarget{hrefBV}{\textbf{BoardVector}} $\mathbf{b}$ is a vector of $K$ non-negative integer numbers $b_k  \in \{0,\ldots,N_k-1\}$. Each $k$ represents a board cell and every board cell $k$ has a predefined number $N_k$ of position values.\footnote{In GBG package \texttt{ntuple2} (base for agent TDNTuple3Agt), all $N_k$ have to be the same. In package \texttt{ntuple4}( base for agent TDNTuple4Agt), numbers $N_k$ may be different for different $k$.} 

A \BoardVector is useful to calculate the feature occurence vector $\mathbf{\Phi}(s)$ in Eq.~\eqref{eq:valueNtuple} for a given n-tuple set:  If an n-tuple contains board cell $k$, then look into $b_k$ to get the position value for this cell $k$. Set $\mathbf{\Phi}_i(s)=1$ for that index $i$ that indexes this n-tuple cell and this position value.

In the following we present different options for boards and \BoardVectors. We do this mainly for the 2x2x2 cube, because it is somewhat simpler to explain. But the same ideas apply to the 3x3x3 cube as well, they are just a little bit longer. Therefore, we defer the lengthy details of the 3x3x3 cube to Appendix~\ref{app:represent-ntuple3x3}.

\subsection{CUBESTATE}
\label{sec:cubestate}

A natural way to translate the cube state into a board is to use the flattened representation of Fig.~\ref{fig:2x2STICKER} as the board and extract from it the 24-element vector $\mathbf{b}$, according to the given numbering. The $k$th element $b_k$ represents a certain cubie face location and gets a number from $\{0,\ldots,5\}$ according to its current face color $f_c$. 
The solved cube is for example represented by $\mathbf{b} = [0000\ 1111\ 2222\ \ldots\ 5555]$. 

This representation CUBESTATE is what the BoardVecType CUBESTATE in our GBG-implementation means: Each board vector is a copy of \texttt{fcol}, the face colors of all cubie faces. \texttt{fcol} is also the vector that uniquely defines each cube state.
An upper bound of possible combinations for $\mathbf{b}$ is $6^{24} = 4.7\cdot 10^{18}$. If we factor out the \DRB-cubie, which always stays at its home position, we can reduce this to 21 board cells with 6 positional values, leading to $6^{21} = \mathbf{2.1\cdot 10^{16}}$ weights. Both numbers are of course way larger than the true number of distinct states (Sec.~\ref{sec:facts2x2}) which is $3.6\cdot 10^{6}$. This is because most of the combinations are dead weights in the n-tuple LUTs, they will never be activated during game play.  

The dead weights occur because many combinations are not realizable, e.g. three white faces in one cubie or any of the $6^3 - 8\cdot 3 = 192$ cubie-face-color combinations that are not present in the real cube. The problem is that the dead weights are scattered in a complicated way among the active weights and it is thus not easy to factor them out.

\begin{figure}%
\centerline{
\begin{tabular}{cc}
	\includegraphics[width=0.25\columnwidth]{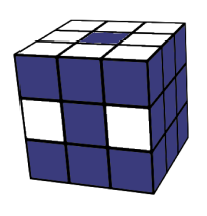}    &
	\includegraphics[width=0.25\columnwidth]{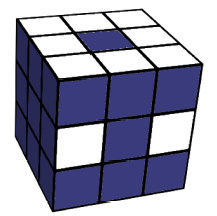} 	\\
	(a) Top view & (b) Bottom view \\
\end{tabular}
}
\caption{The sticker representation used to reduce dimensionality: Stickers that are used are shown in white, whereas ignored stickers are dark blue (from~\cite{mcaleer2019solving}).}%
\label{fig:stickerCube}%
\end{figure}

\begin{figure}%
\centerline{
\renewcommand{\arraystretch}{1.75}
\begin{scriptsize}
\begin{tabular}{|x|x|c|c|x|x|x|x|} \cline{3-4}
\multicolumn{2}{c|}{ } & 3   & 2   &\multicolumn{4}{c}{ } \\ \cline{3-4}
\multicolumn{2}{c|}{ } & 0   & 1   &\multicolumn{4}{c}{ } \\ \hline
                 5 & 4 & 8\bl&11\bl& 18& 17& 23& 22\\  \hline
                 6 & 7 & 9\bl&10\bl& 19& 16& 20& 21\\  \hline
\multicolumn{2}{c|}{ } &14   &13   &\multicolumn{4}{c}{ } \\ \cline{3-4}
\multicolumn{2}{c|}{ } &15   &12\bl&\multicolumn{4}{c}{ } \\  \cline{3-4}
\end{tabular}
\end{scriptsize}
\renewcommand{\arraystretch}{1.0}
}
\caption{Tracked stickers for the 2x2x2 cube (white), while ignored stickers are blue.}
\label{fig:2x2STICKER}
\end{figure}

\subsection{STICKER}
\label{sec:sticker}
\cite{mcaleer2019solving} had the interesting idea for the 3x3x3 cube that 20 \Stickers (cubie faces) are enough. To characterize the full 3x3x3 cube, we need only one (not 2 or 3) sticker for every of the 20 cubies, as shown in Fig.~\ref{fig:stickerCube}. This is because the location of one sticker uniquely defines the location and orientation of that cubie. We name this representation STICKER in GBG.

Translated to the 2x2x2 cube, this means that 8 stickers are enough because we have only 8 cubies. We may for example track the 4 top stickers 0,1,2,3 plus the 4 bottom stickers 12,13,14,15 as shown in Fig.~\ref{fig:2x2STICKER} and ignore the 16 other stickers.
Since we always \href{\#hrefNormalize2x2}{normalize} the cube such that the \DRB-cubie with sticker 12 stays in place, we can reduce this even more to \textbf{7 stickers} (all but sticker 12).

How to lay out this representation as a board? – \cite{mcaleer2019solving} create a rectangular one-hot-encoding board with $7 \times 21 = 147$ cells (7 rows for the stickers and 21 columns for the locations) carrying only 0's and 1's. This is fine for the approach of \cite{mcaleer2019solving}, where they use this board as input for a DNN, but not so nice for n-tuples. Without constraints, such a board amounts to $2^{147} = \mathbf{1.7\cdot 10^{44}}$ combinations, which is unpleasantly large (much larger than in CUBESTATE).\footnote{A possible STICKER \BoardVector for the default cube  would read $\mathbf{b} = [1000000\ 0100000\ 0010000\ \ldots\ ]$, meaning that location 0 has the first sticker, location 1 has the second sticker, and so on. In any STICKER \BoardVector there are only 7 columns carrying exactly one 1, the other carry only 0's. Every row carries exactly one 1.} 

STICKER has more dead weights than CUBESTATE, so it seems like a step back. But the point is, that the dead weights are better structured: If for example sticker 0 appears at column 1 then this column and the two other columns for the same cubie are automatically forbidden for all other stickers. Likewise, if sticker 1 is placed in another column, another set of 3 columns is forbidden, and so on. We can use this fact to form a much more compact representation STICKER2.

\subsection{STICKER2}
\label{sec:sticker2}
As the analysis in the preceding section has shown, the 21 location columns of STICKER cannot carry the tracked stickers in arbitrary combinations. Each cubie (represented by 3 columns in STICKER) carries only exactly \textit{one} sticker. We can make this fact explicit by choosing another representation for the 21 locations: 
$$    \mbox{corner location} = (\mbox{corner cubie}, \mbox{\faceID} ).$$
That is, each location is represented by a pair: corner cubie a,b,c,d,f,g,h (we number the top cubies with letters a,b,c,d and the bottom cubies with letters e,f,g,h and omit e because it corresponds to the \DRB-cubie) and a face ID. To number the faces with a \hypertarget{hrefFaceID}{\textbf{face ID}}, we follow the convention that we start at the top (bottom) face with face ID 1 and then move counter-clockwise around the corner cubie to visit the other faces (2,3). Table~\ref{tab:STICKER2-corner} shows the explicit numbering in this new representation. 

\begin{table}[tbp]%
\caption{The correspondence \textit{corner location $\leftrightarrow$ STICKER2}  for the solved cube. The yellow colored cells show the location of the 7  (2x2x2) and 8 (3x3x3) corner stickers that we track.}
\label{tab:STICKER2-corner}
\centerline{
\begin{scriptsize}
\tabcolsep=0.08cm			
\renewcommand{\arraystretch}{1.1}
\begin{tabular}{|l|c||a|a|a|a||c|c|c|c||a|a|a|a||c|c|c|c||a|a|a|a||c|c|c|c||} \hline
2x2x2		&	location&0\ye&1\ye&2\ye&3\ye &4	&5 &6 &7	 &8	&9	&10	&11	 &12   &13\ye&14\ye&15\ye  &16 &17	&18	&19	 &20	&21	&22	&23 \\ \hline
3x3x3   &	location&0\ye&2\ye&4\ye&6\ye &8	&10&12&14  &16&18 &20	&22	 &24\ye&26\ye&28\ye&30\ye	 &32 &34	&36	&38	 &40	&42	&44	&46 \\ \hline\hline
\multirow{2}{*}{STICKER2 }
        &	corner 	&a   &b  	&c	 &d	   &a &d &h &g   &a &g	&f	&b   &e    &f    &g    &h      &e  &c 	&b 	&f   &e 	&h 	&d 	&c  \\ 
      	&	\faceID &1   &1	  &1   &1    &2	&3 &2 &3	 &3	&2	&3 	&2   &1    &1    &1	   &1	     &2  &2 	&3 	&2 	 &3 	&3 	&2	&3  \\ \hline\hline
\end{tabular}
\renewcommand{\arraystretch}{1.0}
\end{scriptsize}
}
\end{table}

To represent a state as board vector we use now a much smaller board shown in Table~\ref{tab:board-STICKER2}: Each cell in the first row has 7 position values (the letters) and each cell in the second row has 3 position values (the \faceIDs). We show in Table~\ref{tab:board-STICKER2} the board vector  for the default cube, $\mathbf{b} = [\mbox{abcdfgh 1111111}]$. Representation STICKER2 allows for $7^7\cdot 3^7 = \mathbf{1.8\cdot 10^9}$ combinations in total, which is much smaller than STICKER and CUBESTATE.

\begin{table}[ht]
\caption{STICKER2 board representation for the default 2x2x2 cube. For the \BoardVector, cells are numbered row-by-row from 0 to 16.}
\label{tab:board-STICKER2}
\renewcommand{\arraystretch}{1.35}
\centerline{
\begin{tabular}{c|a|a|a|a|a|a|a|c} \cline{2-8}
corner & a\re  & b\re & c\re & d\re & f\re & g\re & h\re & {\scriptsize 7 positions}\\ \cline{2-8}
\faceID& 1     & 1    & 1    & 1    & 1    & 1    & 1    & {\scriptsize 3 positions}\\ \cline{2-8}
\end{tabular}
}
\renewcommand{\arraystretch}{1.0}
\end{table}

STICKER2 has some dead weights remaining, because the combinations can carry the same letter multiple times, which is not allowed for a real cube state. But this rate of dead weights is tolerable. 

It turns out that STICKER2 is in all aspects better than CUBESTATE or STICKER. Therefore, we will only report the results for STICKER2 in the following.

\subsection{Adjacency Sets}
\label{sec:adjacency}

To create n-tuples by random walk, we need adjacency sets (sets of neighbors) to be defined for every board cell $k$. 

For CUBESTATE, the board is the flattened representation of the 2x2x2 cube (Fig.~\ref{fig:2x2x2stickers}). The adjacency set is defined as the 4-point neighborhood, where two stickers are neighbors if they share a common edge on the cube, i.e. are neighbors on the cube.  

For STICKER2, the board consists of 16 cells shown in Table~\ref{tab:board-STICKER2}. Here, the adjacency set for cell $k$ contains all other cells  different from $k$.

\vspace{0.3cm}
Again, the details of ideas similar to Sec.~\ref{sec:cubestate}--\ref{sec:adjacency}, but now for the 3x3x3 cube, are shown in Appendix \ref{sec:cubestate-3x3}--\ref{sec:adjacency3x3}.

\section{Learning the Cube}
\label{sec:learning}

\subsection{McAleer and Agostinelli}
\label{sec:McAleer}

The works of \cite{mcaleer2018solving,mcaleer2019solving} and \cite{agostinelli2019solving} contain up to now the most advanced methods for learning to solve the cube from scratch. 
\cite{agostinelli2019solving} introduces the cost-to-go function for a general Marko decision process
\begin{equation}
			J(s) = \min_{a\in A(s)} \sum_{s'}{P^a(s,s')\left( g^a(s,s')+\gamma J(s') \right)}
\label{eq:Jcost-general}
\end{equation}
where $P^a(s,s')$ is the probability of transitioning from state $s$ to $s'$ by taking action $a$ and $g^a(s,s')$ is the cost for this transition. In the Rubik's cube case, we have deterministic transitions, that is $s'=f(s,a)$ is deterministically prescribed by a forward model $f$. Therefore, the sum reduces to one term and we specialize to $\gamma=1$. Furthermore, we set $g^a(s,s')=1$, because only the length of the solution path counts, so that we get the simpler equation
\begin{equation}
			J(s) = \min_{a\in A(s)} \left( 1+ J(s') \right)  \quad\mbox{with}\quad s'=f(s,a). 
\label{eq:Jcost-rubiks}
\end{equation}

Here, $A(s)$ is the set of available actions in state $s$. We additionally set $J(s^*)=0$ if $s^*$ is the solved cube.
To better understand Eq.~\eqref{eq:Jcost-rubiks} we look at a few examples: If $s_1$ is a state one twist away from $s^*$, Eq.~\eqref{eq:Jcost-rubiks} will find this twist and set $J(s_1)=1$. If $s_2$ is a state two twists away from $s^*$ and all one-twist states have already their correct labels $J(s_1)=1$, then Eq.~\eqref{eq:Jcost-rubiks} will find the twist leading to a $s_1$ state and set $J(s_2)=1+1=2$. While iterations proceed, more and more states (being further away from $s^*$) will be correctly labeled, once their preceding states are correctly labeled. In the end we should ideally have 
		$$ J(s_n)=n. $$

However, the number of states for Rubik's cube is too large to store them all in tabular form. Therefore, \cite{mcaleer2019solving} and \cite{agostinelli2019solving} approximate $J(s)$ with a deep neural network (DNN). To train such a network in the Rubik's cube case, they introduce \hypertarget{hrefDAVI}{\textbf{Deep Approximate Value Iteration (DAVI)}}\footnote{More precisely, \cite{mcaleer2019solving} use Autodidactic Iteration (ADI), a precursor to DAVI, very similar to DAVI, just a bit more complicated to explain. Therefore, we describe here only DAVI.} shown in Algorithm~\ref{algo:DAVI}. The network output $j_{\bfTheta}(s)$ is trained in line 8 to approximate the (unknown) cost-to-go $J(s)$ for every state $s=x_i$.  The main trick of DAVI is, as \cite{agostinelli2019solving} write: \glqq For learning to occur, we must train on a state distribution that allows information to propagate from the goal state to all the other states seen during training. Our approach for achieving this is simple: each training state $x_i$ is obtained by randomly scrambling the goal state $k_i$ times, where $k_i$ is uniformly distributed between $1$ and $K$. During training, the cost-to-go function first improves for states that are only one move away from the goal state. The cost-to-go function then improves for states further away as the reward signal is propagated from the goal state to other states through the cost-to-go function.\grqq

\algnewcommand\And{\textbf{and}}
\begin{algorithm}[tbp]
\caption{DAVI algorithm (from \cite{agostinelli2019solving}). Input: $B$: batch size, $K$: maximum number of twists, $M$: training iterations, $C$: how often to check for convergence, $\epsilon$: error threshold. Output: $\bfTheta$, the trained neural network parameters. 
}
\label{algo:DAVI}
\begin{algorithmic}[1]		
\Function{DAVI}{$B,K,M,C,\epsilon$}  
	\State $\bfTheta \leftarrow$ \Call{initializeNetworkParameters}{}  
	\State $\bfTheta_C \leftarrow \bfTheta$ 
	\For{$m = 1, \ldots, M$}   
			\State $X \leftarrow $\Call{generateScrambledStates}{$B,K$}		\Comment{$B$ scrambled cubes}
			\For{$x_i \in X$}
					\State $y_i \leftarrow \min_{a \in A(s)} \left[ 1+ j_{\bfTheta_C}(f(x_i,a)) \right]$	\Comment{cost-to-go function, Eq.~\eqref{eq:Jcost-rubiks}}
			\EndFor
			\State $(\bfTheta, \mbox{loss}) \leftarrow$ 	\Call{train}{$j_{\bfTheta},X,\mathbf{y}$} \Comment{loss = MSE$(j_{\bfTheta}(x_i),y_i)$}
			\If{($m \mod C=0 \And \mbox{loss}<\epsilon$)} 
					\State $\bfTheta_C \leftarrow \bfTheta$ 
			\EndIf 
	\EndFor
	\State \Return $\bfTheta$
\EndFunction		
\end{algorithmic}
\end{algorithm}

\cite{agostinelli2019solving} use in Algorithm~\ref{algo:DAVI} two sets of parameters to train the DNN: the parameters $\bfTheta$ being trained and the parameters $\bfTheta_C$ used to obtain improved estimates of the cost-to-go function. If they did not use this two separate sets, performance often \glqq saturated after a certain point and sometimes became unstable. Updating $\bfTheta_C$ only after the error falls below a threshold $\epsilon$ yields
better, more stable, performance.\grqq\ \citep{agostinelli2019solving} To train the DNN, they used $M=1\,000\,000$ iterations, each with batch size $B=10\,000$. Thus, the trained DNN has seen ten billion cubes ($10^{10}$) during training, which is still only a small subset of the $4.3\cdot 10^{19}$ possible cube states. 

The heuristic function of the trained DNN alone cannot solve 100\% of the cube states. Especially for higher twist numbers $k_i$, an additional solver or search algorithm is needed. This is in the case of \cite{mcaleer2019solving} a Monte Carlo Tree Search (MCTS),  similar to AlphaZero~\citep{silver2017AlphaGoZero},  which uses the DNN as the source for prior probabilities. \cite{agostinelli2019solving} use instead a variant of A$^*$-search, which is found to produce solutions with a shorter path in a shorter runtime than MCTS.

%
%

\begin{algorithm}[tbp]
\caption{TD-n-tuple algorithm for Rubik's cube. Input: $p_{max}$: maximum number of twists, $M$: training iterations, $E_{train}$: maximum episode length during training, $c$: negative cost-to-go, $R_{pos}$: positive reward for reaching the solved cube $s^*$, $\alpha$: learning rate. $j_{\bfTheta}(s)$: n-tuple network value prediction for state $s$. Output: $\bfTheta$, the trained n-tuple network parameters. 
}
\label{algo:TDNTuple4-rubiks}
\begin{algorithmic}[1]			
\Function{TDNTuple}{$p_{max},M,E_{train},c,R_{pos}$}  
	\State $\bfTheta \leftarrow$ \Call{initializeNetworkParameters}{}  
	\For{$m = 1, \ldots, M$}   
		\State $p \sim U(1,\ldots,p_{max})$				\Comment Draw $p$ uniformly random from $\{1,2,\ldots,p_{max}\}$
		\State $s \leftarrow$ \Call{scrambleSolvedCube}{$p$}   \Comment start state
		\For{$k = 1, \ldots, E_{train}$}   
				\State $s_{new} \leftarrow \underset{a\in A(s)}{\arg\max}\, V(s') \quad\mbox{with} \quad s'=f(s,a) \quad\mbox{and} $
				\State $V(s') = c + \left\{ 
						\begin{array}{l}
							R_{pos}      \mbox{\ \ \qquad if \quad} s'=s^* \\
							\,j_{\bfTheta}(s') \mbox{\qquad if \quad} s'\neq s^*
						\end{array} \right.$
				\State Train network $j_{\bfTheta}$ with Eq.~\eqref{eq:theta} to bring $V(s)$ closer to target $T = V(s_{new})$: 
						$$ V(s) \leftarrow V(s)+\alpha (T-V(s)) $$
				\State $s \leftarrow s_{new}$  
				\If{($s=s^*$)} 
						\State break			\Comment break out of $k$-loop 
				\EndIf 
		\EndFor
	\EndFor
	\State \Return $\bfTheta$
\EndFunction		
\end{algorithmic}
\end{algorithm}

\subsection{N-Tuple-based TD Learning}
\label{sec:tdntuple4}

To solve the Rubik's cube in GBG we use an algorithm that is on the one hand inspired by \DAVI, but on the other hand more similar to traditional reinforcement learning schemes like temporal difference (TD) learning. In fact, we want to use in the end the same TD-FARL algorithm~\citep{Konen2021_FARL_arXiv} that we use for all other GBG games. 

We show in Algorithm~\ref{algo:TDNTuple4-rubiks} our method, that we will explain in the following, highlighting also the similarities and dissimilarities to \DAVI.

First of all, instead of minimizing the positive cost-to-go as in \DAVI, we maximize in lines 7-8 a value function $V(s')$ with a negative cost-to-go. This maximization is functionally equivalent, but more similar to the usual TD-learning scheme. The negative cost-to-go, e.g. $c=-0.1$, plays the role of the positive $1$ in Eq.~\eqref{eq:Jcost-rubiks}.

Secondly, we replace the DNN of \DAVI by the simpler-to-train n-tuple network $j_{\bfTheta}$ with STICKER2 representation as described in Sec.~\ref{sec:ntuples} and \ref{sec:represent-ntuple}. That is, each time $j_{\bfTheta}(s')$ is requested, we first calculate for state $s'$ the \BoardVector in STICKER2 representation, then the occurence vector $\mathbf{\Phi}(s')$ and the value function $V(s')$ according to Eq.~\eqref{eq:valueNtuple}.

The central equations for $V(s')$ in Algorithm~\ref{algo:TDNTuple4-rubiks}, lines 7-8, work similar to Eq.~\eqref{eq:Jcost-rubiks} in \DAVI: If $s=s_1$ is a state one twist away from $s^*$, the local search in $\arg\max V(s')$ will find this twist and the training step in line 9 moves $V(s)$ closer to $c+R_{pos}$.\footnote{
It is relevant, that $R_{pos}$ is a \textbf{positive} number, e.g. 1.0 (and not 0, as it was for \DAVI). This is because we start with an initial n-tuple network with all weights set to 0, so the initial response of the network to any state is 0.0. Thus, if $R_{pos}$ were 0, a one-twist state would see all its neighbors (including $s^*$) initially as responding 0.0 and would not learn the right transition to $s^*$. With $R_{pos}=1.0$ it will quickly find $s^*$.}
Likewise, neighbors $s_2$ of $s_1$ will find $s_1$ and thus move $V(s_2)$ closer to $2c+R_{pos}$.
Similar for $s_3,s_4,\ldots$ under the assumption that a 'known' state  is in the neighborhood. We have a clear gradient on the path towards the solved cube $s^*$.  If there are no 'known' states in the neighborhood of $s_n$, we get for $V(s_n )$ what the net maximally estimates for all those neighbors. We pick the neighbor with the highest estimate, wander around randomly until we hit a state with a 'known' neighbor or until we reach the limit $E_{train}$ of too many steps. 

Note that Algorithm~\ref{algo:TDNTuple4-rubiks} is different from \DAVI insofar that it follows the path $s \rightarrow s'\rightarrow \ldots$ as prescribed by the current $V$, which may lead to a state sequence 'wandering in the unknown' until $E_{train}$ is reached. In contrast to that, \DAVI generates many start states $s_0$ drawn from the distribution of training set states  and trains the network just on pairs $(s_0,T)$, i.e. they do just \textbf{one step} on the path. We instead follow the full path, because we want the training method for Rubik's cube to be as similar as possible to the training method for other GBG games.\footnote{We note in passing that we tested the DAVI variant with $E_{train}=1$ for our TD-n-tuple method as well. 
However, we found that this method gave much worse results, so we stick with our GBG method here.
}

Algorithm~\ref{algo:TDNTuple4-rubiks} is basically the same algorithm as GBG uses for other games. The only differences are (i) the cube-specific start state selection borrowed from \DAVI (a 1-twist start state has the same probability as a 10-twist start state) and (ii) the cube-specific reward in line 8 of Algorithm~\ref{algo:TDNTuple4-rubiks} with its negative cost-go-go $c$ which is however a common element of many RL rewards. 

Algorithm~\ref{algo:TDNTuple4-rubiks} currently learns with only one parameter vector $\bfTheta$. However, it could be extended as in \DAVI to two parameter vectors $\bfTheta$ and $\bfTheta_C$. The weight training step in line 9 is done with the help of Eq.~\eqref{eq:theta} for $\bfTheta$ using the error signal $\delta$ of Eq.~\eqref{eq:TDdelta}.\\[0.1cm]

There are two extra elements, TCL and MCTS, that complete our n-tuple-based TD learning. They are described in the next two subsections.

\subsubsection{Temporal Coherence Learning (TCL)} 
\label{sec:tcl}
The TCL algorithm developed by Beal and Smith~\cite{Beal99} is an extension of TD learning. It replaces the global learning rate $\alpha$ with the weight-individual product $\alpha\alpha_i$ for every weight $\theta_i$. Here, the adjustable learning rate $\alpha_i$ is a free parameter set by a pretty simple procedure: For each weight $\theta_i$, two counters $N_i$ and $A_i$ accumulate the sum of weight changes and the sum of absolute weight changes. If all weight changes have the same sign, then $\alpha_i=|N_i|/A_i=1$, and the learning rate stays at its upper bound. If weight changes have alternating signs, then the global learning rate is probably too large. In this case, $\alpha_i=|N_i|/A_i \rightarrow 0$ for $t \rightarrow \infty$, and the effective learning rate will be largely reduced for this weight. 

In our previous work~\citep{Bagh15} we extended TCL to $\alpha_i=g(|N_i|/A_i)$ where $g$ is a transfer function being either the identity function (standard TCL) or an exponential function $g(x)=e^{\beta(x-1)}$.
It was shown in~\cite{Bagh15} that TCL with this exponential transfer function leads to faster learning and higher win rates for the game ConnectFour.  

\subsubsection{MCTS} 
\label{sec:mcts}

We use Monte Carlo Tree Search (MCTS) \citep{browne2012MCTS} to augment our trained network during testing and evaluation. This is the method also used by \cite{mcaleer2019solving} and by AlphaGo Zero \citep{silver2017AlphaGoZero}, but they use it also during training. 

MCTS builds iteratively a search tree starting with a tree containing only the start state $s_0$ as the root node. 
Until the iteration budget is exhausted, MCTS does the following: In every iteration we start from the root node and select actions following the tree policy until we reach a yet unexpanded leaf node $s_{\ell}$.
The tree policy is implemented in our MCTS wrapper according to the UCB formula~\citep{silver2017AlphaGoZero}:

\begin{eqnarray}
  a_{new} &=& \arg\max_{a \in A(s)}\left(\frac{W(s,a)}{N(s,a)}+U(s,a)\right)
  \label{eq:UCB} \\
	U(s,a) &=& c_{puct}P(s,a)\frac{\sqrt{\varepsilon+\sum_{b \in A(s)}{N(s,b)}}}{1+N(s,a)}
	\label{eq:UCB2}
\end{eqnarray}

Here, $W(s,a)$ is the accumulator for all backpropagated values that arrive along branch $a$ of the node that carries state $s$. Likewise, $N(s,a)$ is the visit counter and $P(s,a)$ the prior probability. $A(s)$ is the set of actions available in state $s$. 
$\varepsilon$ is a small positive constant for the special case $\sum_b{N(s,b)}=0$: It guarantees that in this special case the maximum of $U(s,a)$ is given by the maximum of $P(s,a)$. The prior probabilities $P(s,a)$ are obtained by sending the trained network's values of all follow-up states $s'=f(s,a)$ with $a \in A(s)$ through a softmax function (see Sec.~\ref{sec:ntuples}).\footnote{Note that the prior probabilities and the MCTS iteration are only needed at test time, so that we -- different to AlphaZero -- do not need MCTS during self-play training.} 

Once an unexpanded leaf node $s_{\ell}$ is reached, the node is expanded by initializing its accumulators: $W(s,a) = N(s,a)=0$ and $P(s,a)=p_{s'}$ where $p_{s'}$ is the softmax-squashed output $j_{\bfTheta}(s')$ of our n-tuple network for each state $s'=f(s,a)$. The value of the node is the network output of the best state $j_{\bfTheta}(s_{best})= \max_{s'} j_{\bfTheta}(s')$ and this value is backpropagated up the tree.

More details on our MCTS wrapper can be found in \cite{Scheier2022}.

\begin{algorithm}[tbp]
\caption{TD-n-tuple training algorithm. Input: see Algorithm~\ref{algo:TDNTuple4-rubiks}. Output: $\bfTheta$: trained n-tuple network parameters. 
}
\label{algo:TDNTuple4-training}
\begin{algorithmic}[1]			
\Function{TDNTupleTrain}{$p_{max},M,E_{train},c,R_{pos}$}  
	\State $\bfTheta \leftarrow$ \Call{initializeNetworkParameters}{}  
	\State  \Call{initializeTCLParameters}{}  		\Comment Set TCL-accumulators $N_i=A_i=0, \alpha_i=1 \,\,\forall i$
	\For{$m = 1, \ldots, M$}   
		\State Perform one $m$-iteration of Algorithm~\ref{algo:TDNTuple4-rubiks} with learning rates $\alpha\alpha_i$ instead of $\alpha$ 
		\State $N_i \leftarrow N_i + \Delta \theta_i$ and $A_i \leftarrow A_i + |\Delta \theta_i|$ \Comment Update TCL-accumulators
		\State \Comment where $\Delta \theta_i$ is the last term in Eq.~\eqref{eq:theta}
		\State $\alpha_i \leftarrow |N_i|/A_i  \quad\forall i \mbox{ with } A_i \neq 0$
	\EndFor
	\State \Return $\bfTheta$
\EndFunction		
\end{algorithmic}
\end{algorithm}

\begin{algorithm}[tbp]
\caption{Evaluation algorithm with MCTS solver. Input: trained n-tuple network $j_{\bfTheta}$, $p$: number of scrambling twists, $B$: batch size, $E_{eval}$: maximum episode length during evaluation, $I$: number of MCTS-iterations, $c_{PUCT}$: relative weight for $U(s,a)$ in Eq.~\eqref{eq:UCB}, $d_{max}$: maximum MCTS tree depth. Output: solved rate. 
}
\label{algo:TDNTuple4-evaluation}
\begin{algorithmic}[1]			
\Function{TDNTupleEval}{$j_{\bfTheta},p,B,E_{eval},I,c_{PUCT},d_{max}$}  
	\State $X \leftarrow $\Call{generateScrambledCubes}{$B,p$}		\Comment{$B$ scrambled cubes}
	\State $C_{solved} \leftarrow$ 0
	\For{$x_i \in X$}
		\State $s \leftarrow x_i$				
		\For{$k = 1, \ldots, E_{eval}$}   
				\State $T \leftarrow$ \Call{performMctsSearch}{$s,I,c_{PUCT},d_{max},j_{\bfTheta}$}   
				\State $a \leftarrow$ \Call{selectMostVisitedAction}{}
				\State $s \leftarrow f(s,a) $
				\If{($s=s^*$)} 
						\State $C_{solved} \leftarrow C_{solved}+1$
						\State break			\Comment break out of $k$-loop 
				\EndIf 
		\EndFor
	\EndFor
	\State \Return $C_{solved}/B$		\Comment percentage solved
\EndFunction		
\end{algorithmic}
\end{algorithm}

\subsubsection{Method Summary} 
\label{sec:algo-summary}

We summarize the different ingredients of our n-tuple-based TD learning method in Algorithm~\ref{algo:TDNTuple4-training} (training) and Algorithm~\ref{algo:TDNTuple4-evaluation} (evaluation).

In line 5 of Algorithm~\ref{algo:TDNTuple4-training} we perform one $m$-iteration of Algorithm~\ref{algo:TDNTuple4-rubiks} which does an update step for weight vector $\bfTheta$, see Eq.~\eqref{eq:theta}. All weights of activated n-tuple entries get a weight change $\Delta \theta_i$ equal to the last term in Eq.~\eqref{eq:theta} where the global $\alpha$ is replaced by $\alpha\alpha_i$. 

Line 2 in Algorithm~\ref{algo:TDNTuple4-evaluation} generates a set $X$ of $B$ scrambled cube states. Line 7 builds for each $x_i \in X$ an MCTS tree (see Sec.~\ref{sec:mcts}) starting from root node $x_i$ and line 8 selects the most visited action of the root node. If the goal state $s^*$ is not found during $E_{eval}$ $k$-loop trials, this $x_i$ is considered as not being solved. 

\section{Results}
\label{sec:results}

\subsection{Experimental setup}

We use for all our GBG experiments the same RL method based on n-tuple systems and TCL. Only its hyperparameters are tuned to the specific game, as shown below. We refer to this method/agent as \textbf{\TCLbase} whenever it alone is used  for game playing. If we wrap such an agent by an MCTS wrapper with a given number of iterations, then we refer to this as \textbf{\TCLwrap}.  

We investigate two variants of Rubik's Cube: 2x2x2 and 3x3x3. We trained all TCL agents by presenting them $M=3\,000\,000$ cubes scrambled with $p$ random twists, where $p$ is chosen uniformly at random from $\{1,\ldots,p_{max}\}$. Here, $p_{max}=13\,[16]$ for 2x2x2 and $p_{max}=9\,[13]$ for 3x3x3, where the first number is for \HTM, while the second number in square brackets is for \QTM.  With these $p_{max}$ cube twists we cover the complete cube space for 2x2x2, where God's number (Sec.~\ref{sec:facts}) is known to be $11\,[14]$. But we cover only a small subset in the 3x3x3 case, where God's number is known to be $20\,[26]$~\citep{rokicki2014diameter}.\footnote{We limit ourselves to $p_{max}=9\,[13]$ in the 3x3x3 \HTM [\QTM] case, because our network has not enough capacity to learn all states of the 3x3x3 Rubik’s cube. Experiments with higher twist numbers during training did not improve the solved-rates.} We train 3 agents for each cube variant \{ 2x2x2, 3x3x3 \} $\times$ \{ HTM, QTM \} to assess the variability of training.

The hyperparameters of the agent for each cube variant were found by manual fine-tuning. 
For brevity, we defer the exact explanation and setting of all parameters to Appendix~\ref{app:hyperparams}.

We evaluate the trained agents for each $p$ on 200 scrambled cubes that are created by applying the given number $p$ of random scrambling twists to a solved cube. The agent now tries to solve each scrambled cube. A cube is said to be \textit{unsolved} during evaluation if the agent cannot reach the solved cube in $E_{eval}=50$ steps.\footnote{During training, we use lower maximum episode lengths $E_{train}$ (see Appendix~\ref{app:hyperparams}) than $E_{eval}=50$ in order to reduce computation time (in the beginning, many episodes cannot be solved, and $50$ would waste a lot of computation time). But $E_{train}$ is always at least $p_{max}+3$ in order to ensure that the agent has a fair chance to solve the cube and collect the reward.}

\begin{figure}[tbp]%
\centerline{
	\includegraphics[width=0.5\columnwidth]{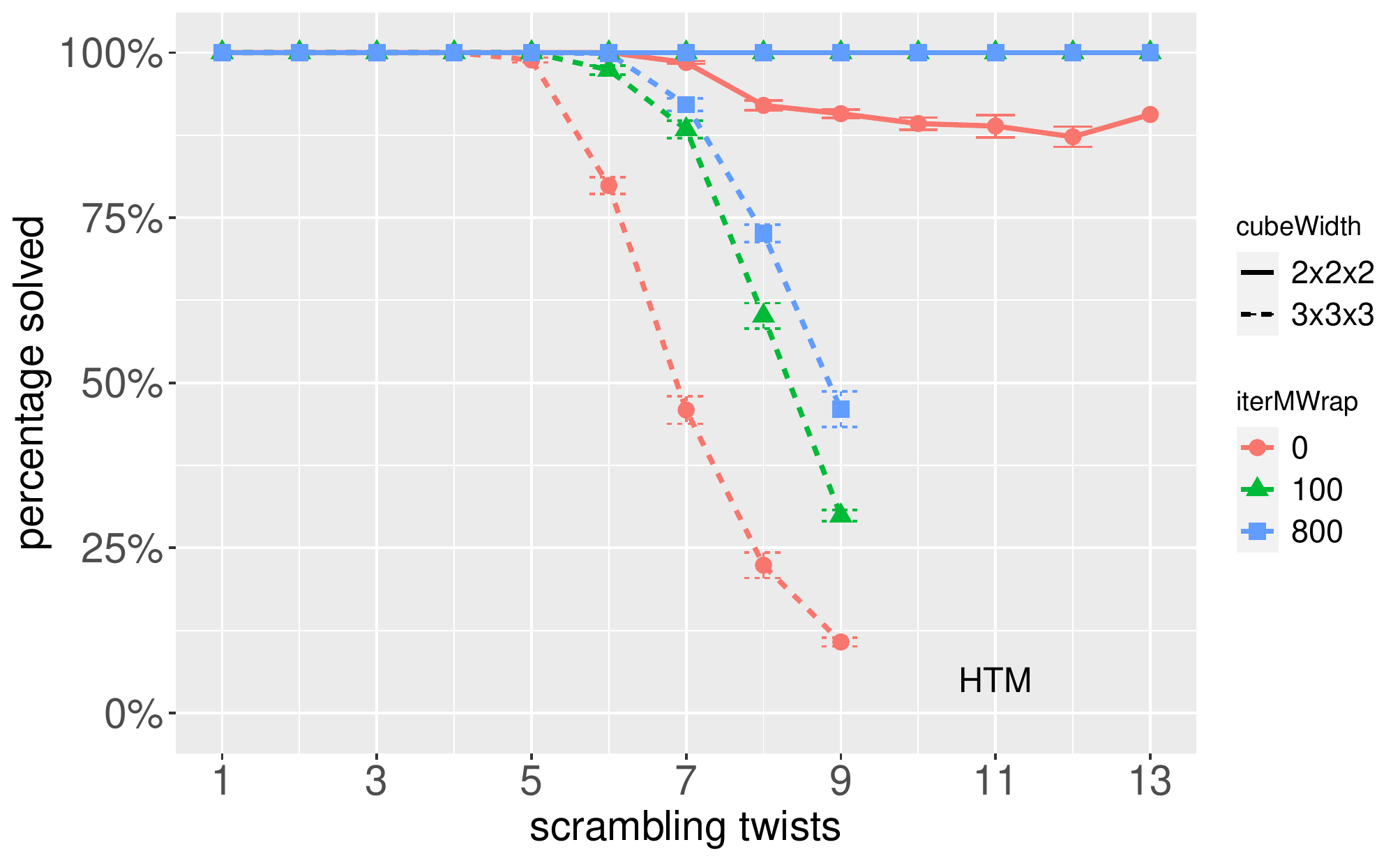}    
	\includegraphics[width=0.5\columnwidth]{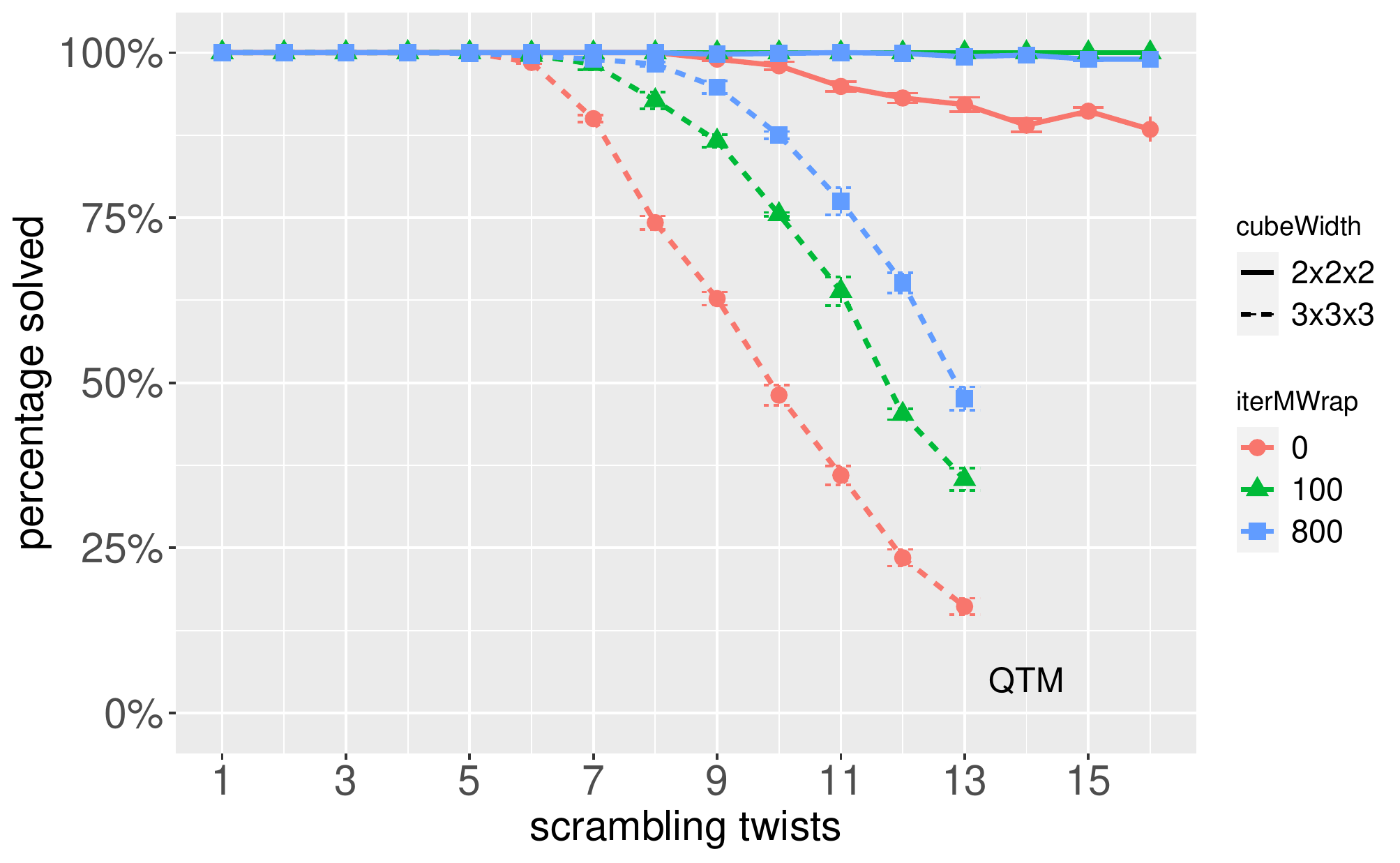}    
}
\caption{Percentage of solved cubes as a function of scrambling twists $p$ for the trained TD-N-tuple agent wrapped by MCTS wrapper with different numbers of iterations. The red curves are \TCLbase without wrapper, the other colors show different forms of \TCLwrap. Twist type is \HTM (left) and \QTM (right). Each point is the average of 3 independently trained agents.}%
\label{fig:solvedRate-ptwist}%
\end{figure}

\subsection{Cube Solving with MCTS Wrapper, without Symmetries}
\label{sec:resWrapper}

The trained TD-N-tuple agents learn to solve the cubes to some extent, as the red curves \TCLbase in Fig.~\ref{fig:solvedRate-ptwist} show, but they are in many cases (i.e. $p>p_{max}/2$) far from being perfect. These are the results from training each agent for 3 million episodes, but the results would not change considerably, if 10 million training episodes were used.

\cite{Scheier2022} have shown, that the performance of agents, namely TD-N-tuple agents, is largely improved, if the trained agents are wrapped during test, play and evaluation by an MCTS wrapper. This holds for Rubik's cube as well, as Fig.~\ref{fig:solvedRate-ptwist} shows: For the 2x2x2 cube, the non-wrapped agent \TCLbase (red curve) is already quite good, but with wrapping it becomes almost perfect. For the 3x3x3 cube, the red curves are not satisfactorily: the solved-rates are below 20\% for $p=9\, [13]$ in the \HTM  [\QTM] case. But at least MCTS wrapping boosts the solved-rates by a factor of 3 [QTM: from 16\% to 48\%] or 4.5 [HTM: from 10\% to 45\%].

All these results are without incorporating symmetries. How symmetries affect the solved-rates will be investigated in Sec.~\ref{sec:resSymmetry}. But before this, we look in the next section at the number of symmetries that effectively exist in a cube state.

\begin{figure}%
\centerline{
	\includegraphics[width=\columnwidth]{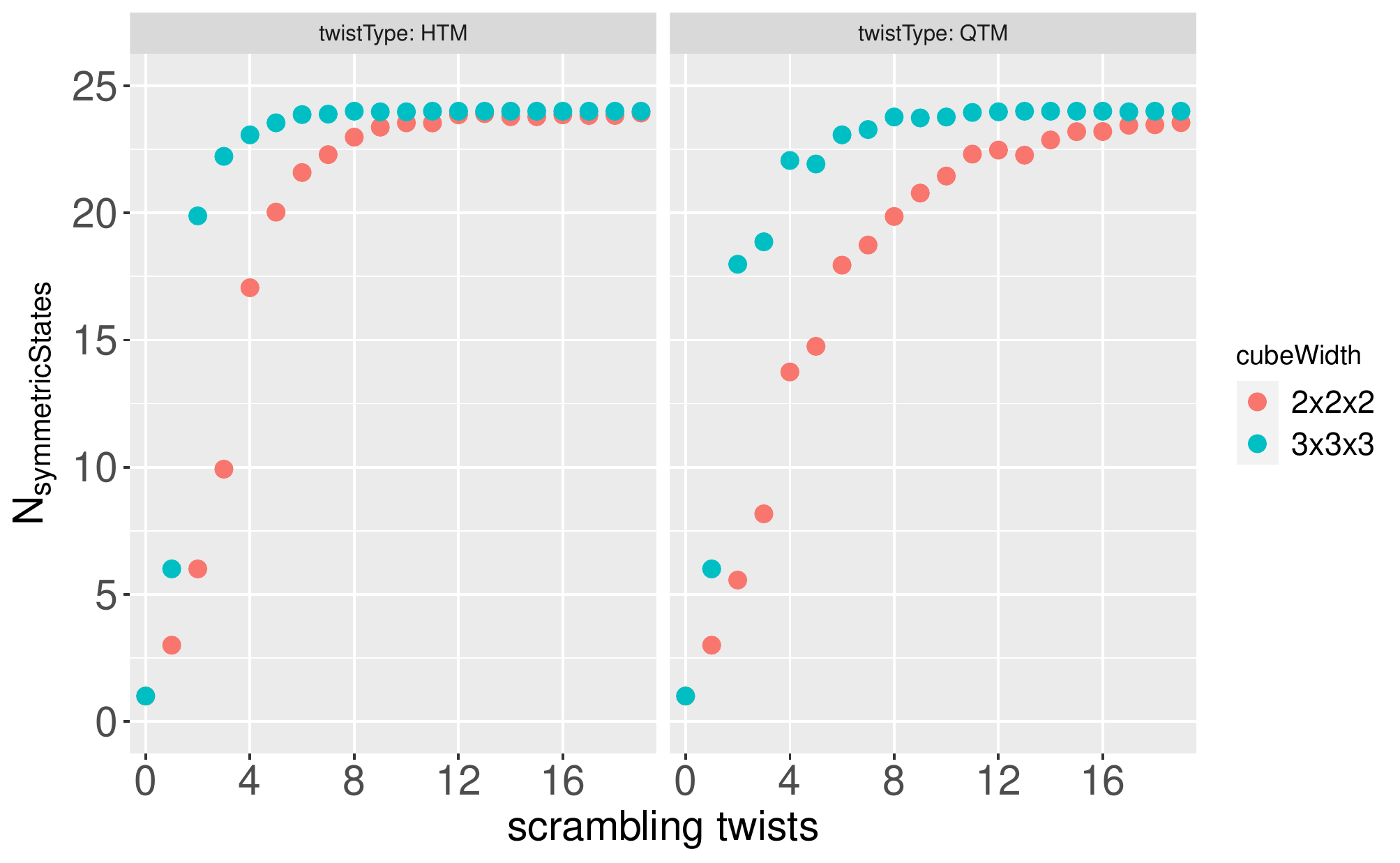}    
}
\caption{Count of truly different symmetric states for cube states generated by $p$ random scrambling twists. Each point is an average over 500 such states.}%
\label{fig:numSymStates}%
\end{figure}

\subsection{Number of Symmetric States}
\label{sec:numSymmetry}
Not every cube state has 24 truly different symmetric states (24 = number of color symmetries). For example in the solved cube, all color-symmetric states are the same (after normalization). Thus, we have here only one truly different symmetric state. 

However, we show in this section that for the majority of cube states the number of truly different symmetric states is close to 24. Two states are truly different if they are not the same after the \href{\#hrefNormalize2x2}{normalizing operation}. We generate a cube state by applying $p$ random scrambling twists to the default cube.
Now we apply all 24 color transformations (Sec.~\ref{sec:colortrans}) to it and count the truly different states. The results are shown in Fig.~\ref{fig:numSymStates} for both cube sizes and both twist types.  For the 3x3x3 cube, the number of states quickly (for $p>5$) approaches the maximum $N=24$, while for the 2x2x2 cube it is a bit slower: $p>4$ or $p>8$ is needed to surpass $N=20$. 

As a consequence, it makes sense to use 16 or even 24 symmetries when training and evaluating cube agents. Especially for scrambled states with higher $p$, the 24 color transformations used to construct symmetric states will usually lead to 24 different states.

\begin{figure}[tbp]%
\centerline{
	\includegraphics[width=0.8\columnwidth]{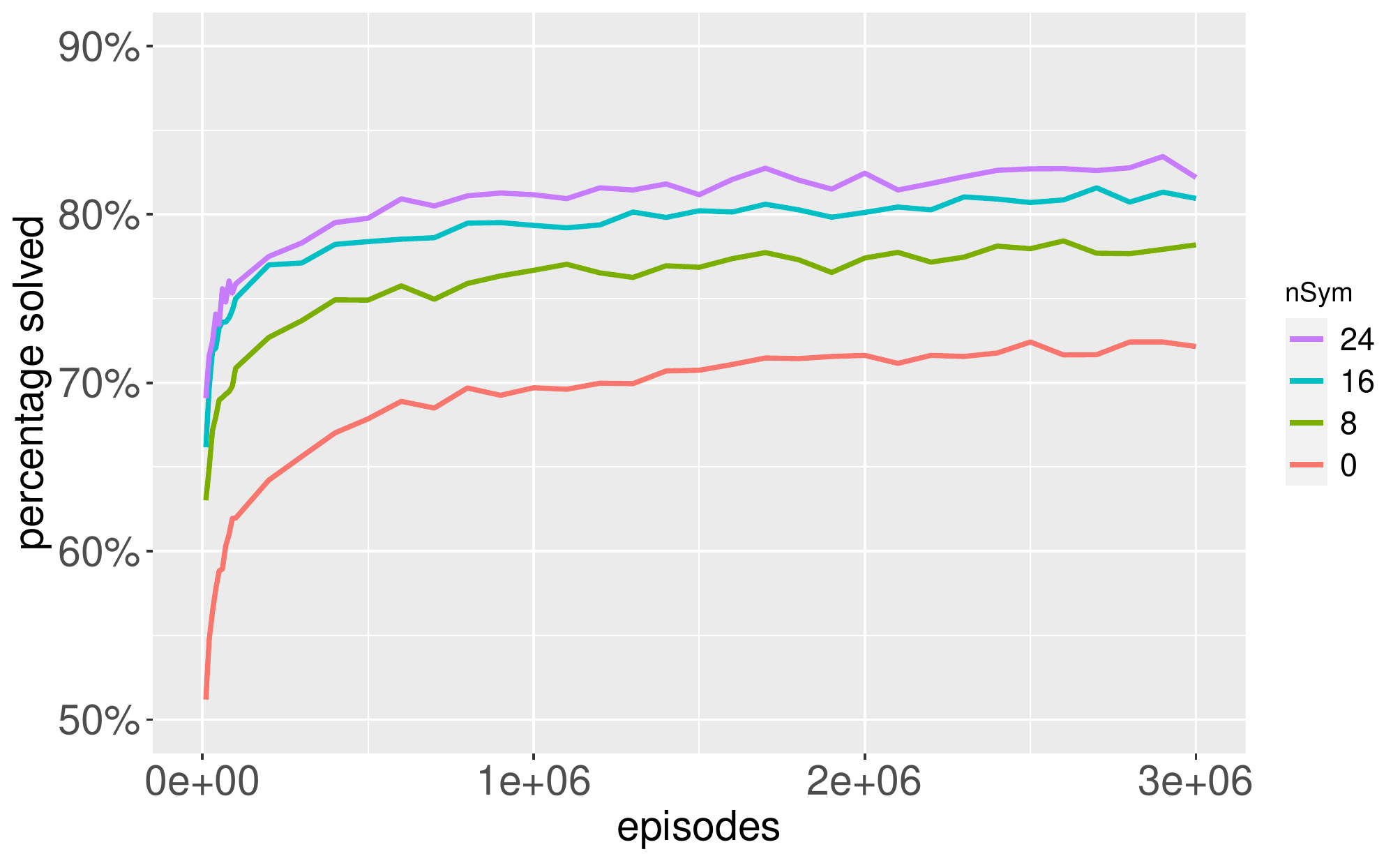}    
}
\caption{Learning curves for different numbers \texttt{nSym} $=0,8,16,24$ of symmetries. Shown is the solved rate of (3x3x3, QTM) cubes. The solved rate is the average over all twist numbers $p=1,\ldots,13$ with 200 testing cubes for each $p$ and over 3 agents with different random-walk n-tuple sets.
}%
\label{fig:rubiks-learncurves}%
\end{figure}

\subsection{The Benefit of Symmetries}
\label{sec:resSymmetry}

\begin{figure}[tbh]%
\centerline{
	\includegraphics[width=0.8\columnwidth]{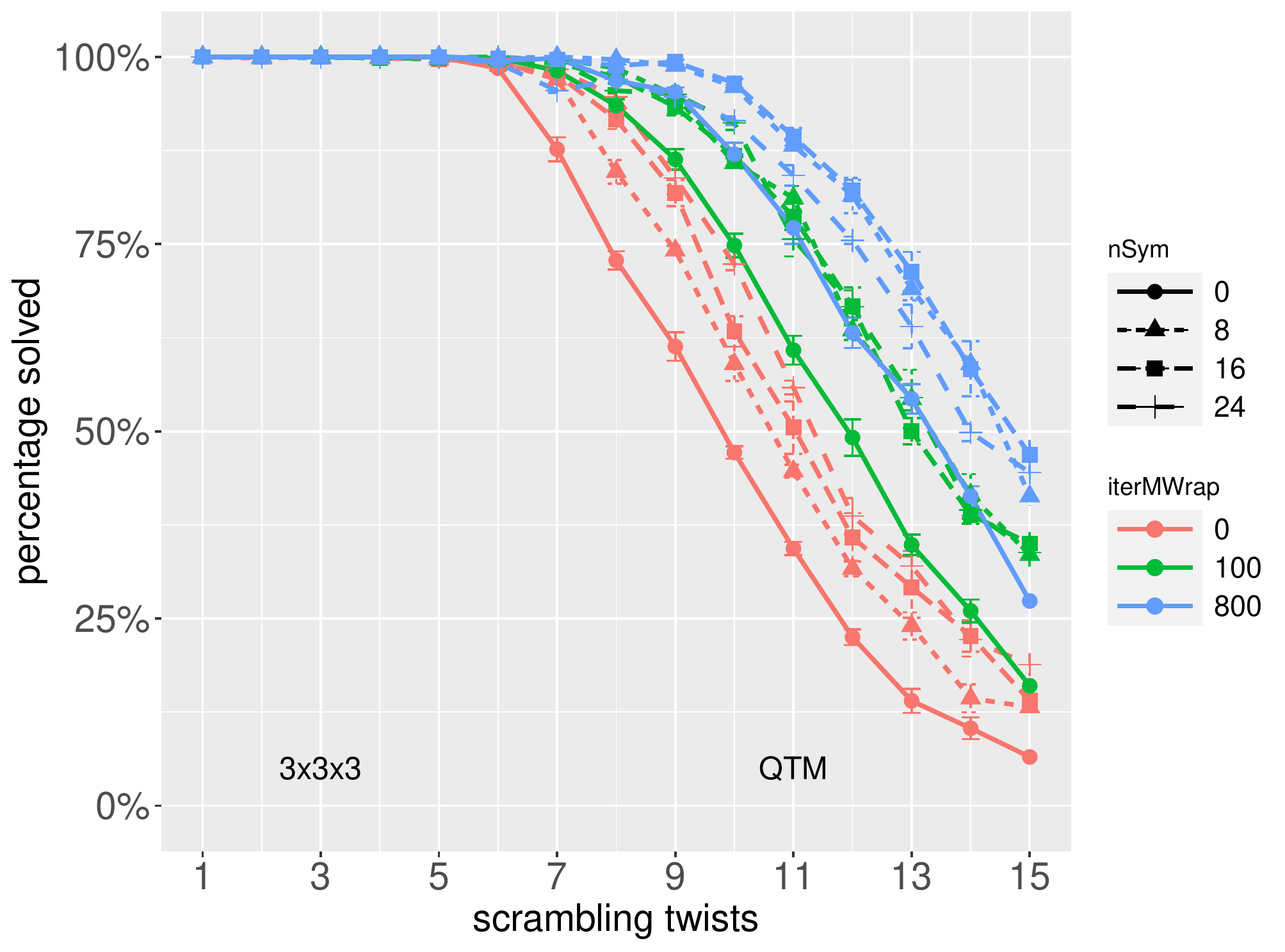}    
}
\caption{With symmetries: Percentage of solved cubes (3x3x3, QTM) as a function of scrambling twists $p$ for TD-N-tuple agents trained and evaluated with different numbers of symmetries \texttt{nSym} and wrapped by MCTS wrappers with different iterations. The red curves are \TCLbase (without wrapper), the other colors show different forms of \TCLwrap. The solved rates are the average over 200 testing cubes for each $p$ and over 3 agents with different random-walk n-tuple sets.
}%
\label{fig:solvedRate-nSym}%
\end{figure}

In order to investigate the benefits of symmetries, we first train a TCL agent with different numbers of symmetries. As described in Sec.~\ref{sec:symmetr}, we select in each step \texttt{nSym} $=0,8,16,24$ symmetric states. Which symmetric states are chosen is selected randomly. Symmetries are used (a) to update the weights for each symmetric state and (b) to build with Eq.~\eqref{eq:Vsym} a smoothed value function which is used to decide about the next action during training. For $0, 8, 16, 24$ symmetries, we train 3 agents each (3x3x3 cube, STICKER2, QTM). The 3 agents differ due to their differently created random-walk n-tuple sets.

Fig.~\ref{fig:rubiks-learncurves} shows the learning curves for different \texttt{nSym} $=0,8,16,24$. It is found that agents with \texttt{nSym} $>0$ learn faster and achieve a higher asymptotic solved rate.

Next, we evaluate each of the trained agents by trying to solve for each $p \in \{1,\ldots,15\}$ (scrambling twists) 200 different scrambled cubes. During evaluation, we use again the same \texttt{nSym} as in training to form a smoothed value function. 
We compare in Fig.~\ref{fig:solvedRate-nSym} different symmetry results, both without wrapping (\TCLbase, red curves) and with MCTS-wrapped agents using 100 (green) or 800 (blue) iterations. It is clearly visible that MCTS wrapping has a large effect, as it was also the case in Fig~\ref{fig:solvedRate-ptwist}. But in addition to that, the use of symmetries leads for each agent, wrapped or not, to a substantial increase in solved-rates (a surplus of 10-20\%). It is remarkable, that even for $p$=14 or 15 a solved rate above or near 50\% can be reached\footnote{$p$ is above $p_{max}$=13, the maximum twist number used during training.} by the combination (\texttt{nSym}=16,  800 MCTS iterations). 

Surprisingly, it seems that \textit{with wrapping} it is only important whether we use symmetries, not how many, since the difference between  \texttt{nSym} $=8,16,24$ is only marginal. For 800 MCTS iterations, the solved rate for \texttt{nSym} $=24$ is in most cases even smaller than that for \texttt{nSym} $=8,16$. This is surprising because it would have been expected that also with wrapping a larger \texttt{nSym} should lead to a smoother value function and thus should in theory produce larger solved rates. -- 
Note that this is not a contradiction to Fig.~\ref{fig:rubiks-learncurves}, because the learning curves were obtained \textit{without wrapping} and the red \TCLbase curves in Fig.~\ref{fig:solvedRate-nSym} (again without wrapping) show the same positive trend with increasing \texttt{nSym}\footnote{i.e. \texttt{nSym}$=24$ is for every $p$ clearly better than \texttt{nSym}$=16$}. The red curves in Fig.~\ref{fig:solvedRate-nSym} show approximately the same average solved rates as the asymptotic values in Fig.~\ref{fig:rubiks-learncurves}. 



\begin{table}[tbp]
\caption{Computation times with symmetries. All numbers are for 3x3x3 cube, STICKER2 and QTM. Training: 3 million self-play episodes, w/o MCTS in the training loop. Testing: 200 scrambled cubes with $p=13$, agents wrapped by MCTS wrapper with \texttt{iter} iterations. }
\label{tab:compTimes}
\centerline{
\begin{tabular}{|c||c||r|r|r|r|r|} \cline{1-7}
\multirow{3}{*}
{\texttt{nSym}} & training & \multicolumn{5}{c|}{ testing } \\
                & [hours]  & \multicolumn{5}{c|}{ [seconds] } \\ \cline{3-7}
								& & \texttt{iter} & 0   &  100  &  400 &  800 \\ \hline\hline
						0   &   0.5    &      & 0.5 &   48  &  196 &  390 \\ \hline		
						8   &   5.4    &      & 4.0 &  241  &  877 & 1400 \\ \hline		
					 16   &   9.5    &      & 7.3 &  464  & 1380 & 2330 \\ \hline		
					 24   &  13.0    &      & 8.0 &  550  & 1760 & 3130 \\ \hline		
\end{tabular}
}
\end{table}

\subsection{Computational Costs}
\label{sec:compTimes}

Table~\ref{tab:compTimes} shows the computational costs when training and testing with symmetries.
All computations were done on a single CPU Intel i7-9850H @ 2.60GHz.
If we subtract the computational costs for \texttt{nsym}$=0$, computation time increases more or less linearly with \texttt{iter} and roughly linearly with \texttt{nSym}. Computation times for \texttt{nSym}$=24$ are approximately 10x larger than those for \texttt{nSym}$=0$.

Computation times are dependent on the solved rate: If a cube with $p=13$ is solved, the episode takes normally 12-15 steps. If the cube is not solved, the episode needs 50 steps, i.e. a factor of 3-4 more. Thus, the numbers in Table~\ref{tab:compTimes} should be taken only as rough indication of the trend.

Bottom line: Training time through symmetries increases by a factor of $13/0.5 = 26$ (\texttt{nSym}$=24$) and testing time increases through 800 MCTS iterations by a factor of about $3130/8 \approx 400$.

Training with symmetries takes between 5.4h and 13h on a normal CPU, depending on the number of symmetries. This is much less than the 44h on a 32-core server with 3 GPUs that were used by \cite{mcaleer2019solving}. But it also does not reach the same quality as \cite{mcaleer2019solving}.

\section{Related Work}
\label{sec:rel-work}

Ernö Rubik invented Rubik's cube in 1974. Rubik's cube has gained worldwide popularity with many human-oriented algorithms being developed to solve the cube from arbitrary scrambled start states. By 'human-oriented' we mean algorithms that are simple to memorize for humans. They usually will find long, suboptimal solutions. For a long time it was an unsolved question what is the minimal number of moves (\href{\#hrefGodsNum}{God's Number}) needed to solve any given cube state. The early work of \cite{thistle1981} put an upper bound on this number with his 52-move algorithm. This was one of the first works to systematically use group theory as an aid to solve Rubik's cube. Later, several authors have gradually reduced the upper bound 52~\citep{joyner2014man}, until \cite{rokicki2014diameter} could prove in 2014 for the 3x3x3 cube that \href{\#hrefGodsNum}{God's Number} is 20 in \HTM and 26 in \QTM.

Computer algorithms to solve Rubik's cube rely often on hand-engineered features and group theory. One popular solver for Rubik's cube is the two-phase algorithm of \cite{kociemba2015two}. A variant of A$^*$ heuristic search was used by \cite{korf1991maxN}, along with a pattern database heuristic, to find the shortest possible solutions.

The problem of letting a computer \textit{learn} to solve Rubik's cube turned out to be much harder: \cite{irpan2016exploring} experimented with different neural net baseline architectures (LSTM gave for him reportedly best results) and tried to boost them with AdaBoost. However, he had only for scrambling twist $\leq 7$ solved rates of better than 50\% and the baseline turned out to be better than the boosted variants. \cite{brunetto2017deep} found somewhat better results with a DNN, they could solve cube states with 18 twists with a rate above 50\%. But they did not learn from scratch because they used an optimal solver based on \cite{kociemba2015two} to generate training examples for the DNN. \cite{smith2016discovering} tried to learn Rubik's cube by genetic programming. However, their learned solver could only reliably solve cubes with up to 5 scrambling twists. 

A breakthrough in learning to solve Rubik's cube are the works of \cite{mcaleer2018solving,mcaleer2019solving} and \cite{agostinelli2019solving}: With Autodidactic Iteration (ADI) and Deep Approximate Value Iteration (\DAVI) they were able \textit{to learn from scratch} to solve Rubik's cube in \QTM for arbitrary scrambling twists. Their method has been explained in detail already in Sec.~\ref{sec:McAleer}, so we highlight here only their important findings: \cite{mcaleer2019solving} only needs to inspect less than 4000 cubes with its trained network DeepCube when solving for a particular cube, while the optimal solver of \cite{korf1991maxN} inspects 122 billion different nodes, so Korf's method is much slower. 

\cite{agostinelli2019solving} extended the work of \cite{mcaleer2019solving} by replacing the MCTS solver with a batch-weighted A$^*$ solver which is found to produce shorter solution paths and have shorter run times. At the same time, \cite{agostinelli2019solving} applied their agent DeepCubeA successfully to other puzzles like LightsOut, Sokoban, and the 15-, 24-, 35- and 48-puzzle\footnote{a set of 15, 24, ... numbers has to be ordered on a $4\times 4$, $5\times 5$, ... square with one empty field}. DeepCubeA could solve all of them.

The deep network used by \cite{mcaleer2019solving} and \cite{agostinelli2019solving} were trained without human knowledge or supervised input from computerized solvers. The network of \cite{mcaleer2019solving} had over 12 million weights and was trained for 44 hours on a 32-core server with 3 GPUs. The network of  \cite{mcaleer2019solving} has seen 8 billion cubes during training. -- Our approach started from scratch as well. It required much less computational effort (e.g. 5.4h training time on a single standard CPU for nSym=8, see Table~\ref{tab:compTimes}).
It can solve the 2x2x2 cube completely, but the 3x3x3 cube only partly (up to 15 scrambling twists). Each trained agent for the 3x3x3 cube has seen 48 million scrambled cubes\footnote{$3\cdot10^6\,\times\,16 =$ training episodes $\times$ episode length $E_{train}$. This is an upper bound: some episodes may have shorter length, but each unsolved episode has length $E_{train}$.} during training.

\section{Summary and Outlook}
\label{sec:summary}

We have presented new work on how to solve Rubik's cube with n-tuple systems, reinforcement learning and an MCTS solver. The main ideas were already presented in \cite{Scheier2022} but only for \HTM and up to $p=9$ twists. Here we extended this work to \QTM as well and presented all the details 
of cube representation and n-tuple learning algorithms necessary to reproduce our Rubik's cube results.
As a new aspect, we added cube symmetries and studied their effect on  solution quality. We found that the use of symmetries boosts the solved rates by 10-20\%. Based on this, we could increase for \QTM the number of scrambling twists where at least 45\% of the cubes are solved from $p=13$ without symmetries to $p=15$ with symmetries. 

We cannot solve the 3x3x3 cube completely, as \cite{mcaleer2019solving} and \cite{agostinelli2019solving} do. But our solution is much less computational demanding than their approach. 

Further work might be to look into larger or differently structured n-tuple systems, perhaps utilizing the staging principle that \cite{jaskowski2018mastering} used to produce world-record results in the game 2048.

\newpage

\newpage
\appendix
\noindent{\LARGE\textbf{Appendix}}

\section{Calculating \texttt{sloc} from \texttt{fcol}}
\label{app:calc_s_from_f}

Given the face colors $f_c$ (Eq.~\eqref{eq:fcol}) of a transformed cube, how can we calculate the transformed sticker locations \sL (Eq.~\eqref{eq:sloc})?

This problem seems ill-posed at first sight, because a certain face color, e.g. \textit{white}, appears multiple times in $f_c$ and it is not possible to tell from the appearance of \textit{white} alone to which sticker location \sL it corresponds. But with a little more effort, i.e. by looking at the neighbors of the \textit{white} sticker, we can solve the problem, as we show in the following.

\subsection{2x2x2 cube}
All \Cubies of the 2x2x2 cube are corner cubies. We track for each cubie exactly one \Sticker. This can be for example the set 
		$$\mathfrak{B}=\{0,1,2,3,12,13,14,15\}$$ 
of 8 stickers, which is the same as the set of tracked stickers shown in Fig.~\ref{fig:2x2STICKER}. 

For each $s \in \mathfrak{B}$:
\begin{enumerate}
	\item Build the cubie that contains $s$ as the first sticker.\footnote{We know for example from looking at the default cube in Fig.~\ref{fig:2x2STICKER} that sticker $s=0$ is part of the $0$-$8$-$4$-cubie.} 
	\item Locate the cubie in $f_c$. That is, find a location in $f_c$ with the same color as the $1^{st}$ cubie face. If found, check if the neighbor to the right\footnote{By \textit{neighbor to the right} we mean the next sticker when we march in clockwise orientation around the actual cubie.} has the color of the $2^{nd}$ cubie face. If yes, check if its neighbor to the right has the color of the $3^{rd}$ cubie face. If yes, we have located the cubie in $f_c$ and we return it, i.e. its three sticker locations $C = [a,b,c]$.
	\item Having located the cubie, we can infer three elements of \sL:
	\begin{eqnarray}
			\sL[s] 			 &=& C[0] \\
			\sL[R[s]] 	 &=& C[1] \\
			\sL[R[R[s]]] &=& C[2]
	\label{eq:sL_locate}
	\end{eqnarray}
	Here $R[s]$ is the right neighbor of sticker $s$. $R[R[s]]]$ is the left neighbor.
\end{enumerate}

In total, we have located $8\times 3 = 24$ stickers, i.e. the whole transformation for \sL.\footnote{
{\scriptsize The relevant GBG source code is in CubeState.locate and CubeState2x2.apply\_sloc\_slow.}}

\subsection{3x3x3 cube}
The 3x3x3 cube has 8 corner \Cubies and 12 edge cubies. We track for each cubie exactly one \Sticker. This can be for the corners the set 
		$$\mathfrak{B}=\{0,2,4,6,24,26,28,30\}$$ 
and for the edges the set 
		$$\mathfrak{E}=\{1,3,5,7, 25,27,29,31, 11,15,21,33\}.$$ 

We do for the corner set $\mathfrak{B}$ the same as we did for the 2x2x2 cube. 

For each element $s \in \mathfrak{E}$ of the edge set:
\begin{enumerate}
	\item Build the edge cubie $c_E$ that contains $s$ as the first sticker.
	\item Locate the cubie in $f_c$. That is, find an edge location in $f_c$ with the same color as the $1^{st}$ cubie face. If found, check if the other sticker of that cubie has the same color as the other sticker of $c_E$. If yes, we have located the edge cubie in $f_c$ and we return it, i.e. its two stickers $C = [a,b]$.
	\item Having located the cubie, we can infer two elements of \sL:
	\begin{eqnarray}
			\sL[s] 			 &=& C[0] \\
			\sL[O[s]] 	 &=& C[1] 
	\label{eq:sL_locate_edge}
	\end{eqnarray}
	Here $O[s]$ is the other sticker of the edge cubie that has sticker $s$ as first sticker.
\end{enumerate}

\noindent
In total, we have located 
		$$8\times 3 + 12\times 2= 48$$ 
stickers, i.e. the whole transformation for \sL.\footnote{
{\scriptsize The relevant GBG source code is in CubeState.locate, CubeState3x3.locate\_edge and CubeState3x3.apply\_sloc\_slow.}}

\section{N-Tuple Representations for the 3x3x3 Cube}
\label{app:represent-ntuple3x3}

In this appendix we describe the n-tuple representations of the cube, analogously to the 2x2x2 cube Sec.~\ref{sec:represent-ntuple}, but now for the 3x3x3 cube.

\subsection{CUBESTATE}
\label{sec:cubestate-3x3}
A natural way to translate the cube state into a board is to use the flattened representation of Fig.~\ref{fig:3x3x3stickers} as the board and extract from it the 48-element vector $\mathbf{b}$, according to the given numbering. The $k$th element $b_k$ represents a certain cubie face location and gets a number from $\{0,\ldots,5\}$ according to its current face color $f_c$. 
The solved cube is for example represented by $\mathbf{b} = [00000000\ 11111111\ \ldots\ 55555555]$. 

This representation CUBESTATE is what the BoardVecType CUBESTATE in our GBG-implementation means: Each board vector is a copy of \texttt{fcol}, the face colors of all cubie faces. An upper bound of possible combinations for $\mathbf{b}$ is $6^{48} = \mathbf{2.2\cdot 10^{32}}$. This is much larger than the true number of distinct states (Sec.~\ref{sec:facts3x3}) which is $4.3\cdot 10^{19}$. 

\subsection{STICKER}
\label{sec:sticker-3x3}
\cite{mcaleer2019solving} had the interesting idea for the 3x3x3 cube that 20 \Stickers (cubie faces) are enough. To characterize the 3x3x3 cube, we need according to \cite{mcaleer2019solving} only one (not 2 or 3) sticker for every of the 20 cubies, as shown in Fig.~\ref{fig:stickerCube}. This is because the location of one sticker uniquely defines the location and orientation of that cubie. We name this representation STICKER in GBG.

We track the 4 top corner stickers 0,2,4,6 plus the 4 bottom corner stickers 24,26,28,30 plus one sticker for each of the 12 edge stickes as shown in Fig.~\ref{fig:stickerCube}, in total 20 stickers and ignore the 28 other stickers. 

How to lay out this representation as a board? – \cite{mcaleer2019solving} create a rectangular one-hot-encoding board with $20 \times 24 = 480$ cells (20 rows for the stickers and 24 columns for the locations\footnote{$8\cdot 3$ for the corner stickers and $12\cdot 2$ for the edge stickers}) carrying only 0's and 1's. This is fine for the approach of \cite{mcaleer2019solving}, where they use this board as input for a DNN, but not so nice for n-tuples. Without constraints, such a board amounts to $2^{480} \approx 10^{145}$ combinations, which is unpleasantly large (much larger than in CUBESTATE).\footnote{\cite{mcaleer2019solving} do not need a weight for every of the $2^{480}$ possible states, as the n-tuple network would need. Instead they need only $480\cdot 4096 = 2\cdot 10^6$ weights to the first hidden layer having 4096 neurons.} 

Another possibility to lay out the board: Specify 20 board cells (the stickers) with 24 position values each. This amounts to $24^{20} = \mathbf{4.0\cdot 10^{27}}$ combinations.


\subsection{STICKER2}
\label{sec:sticker2-3x3}

Analogously to Sec.~\ref{sec:sticker2}, we represent the 24 corner locations and 24 edge locations as: 
$$    \mbox{corner location} = (\mbox{corner cubie}, \mbox{\faceID} ),$$
$$    \mbox{edge location} = (\mbox{edge cubie}, \mbox{\faceID} ).$$
That is, each corner location is represented by a corner cubie a,b,c,d,e,f,g,h and by a \faceID 1,2,3. Table~\ref{tab:STICKER2-corner} shows the explicit numbering in this new representation. Additionally, each edge location is represented by an edge cubie A,B,C,D,E,F,G,H,I,J,K,L\footnote{4 U-stickers, 4 D-sticker, 4 middle-layer stickers (2F, 2B)} and by a face ID 1,2.  Convention for face ID numbering of edge cubies: For top- and bottom-layer edge cubies, it is 1 for U and D stickers, 2 else. The face ID for middle-layer edge cubies is 1 for F and B stickers, 2 else. Table~\ref{tab:STICKER2-edge} shows the explicit numbering in this representation.  

\begin{table}[tbp]%
\caption{The correspondence \textit{edge location $\leftrightarrow$ STICKER2}  for the solved cube. The yellow colored cells show the location of the 12 edge stickers that we track.}
\label{tab:STICKER2-edge}
\centerline{
\begin{scriptsize}
\tabcolsep=0.08cm			
\renewcommand{\arraystretch}{1.1}
\begin{tabular}{|l|c||a|a|a|a||c|c|c|c||a|a|a|a||c|c|c|c||a|a|a|a||c|c|c|c||} \hline
3x3x3   &	location&1\ye&3\ye&5\ye&7\ye &9	&11&13&15  &17\ye&19&21\ye&23	 &25\ye&27\ye&29\ye&31\ye	 &33 &35	&37	&39	 &41	&43\ye&45	&47\ye \\ \hline\hline
\multirow{2}{*}{STICKER2 }
        &	edge   	&A   &B  	&C	 &D	   &D &G &K &E      &E &J	&F	  &A   &I    &J    &K    &L      &H  &B 	&F 	&I   &L 	&G 	  &C 	&H  \\ 
      	&	\faceID &1   &1	  &1   &1    &2	&2 &2 &2	    &1 &1	&1 	  &1   &1    &1    &1	   &1	     &2  &2 	&2 	&2 	 &1 	&1 	  &1	&1  \\ \hline\hline
\end{tabular}
\renewcommand{\arraystretch}{1.0}
\end{scriptsize}
}
\end{table}

The corresponding board consists of 8 + 8 + 12 +12 = 40 cells shown in Table~\ref{tab:board-STICKER2-3x3}. The 8 cell pairs in the first two rows code the locations of the tracked corner stickers 0,2,4,6,24,26,28,30, see Table~\ref{tab:STICKER2-corner} in Sec.~\ref{sec:sticker2}. The 12 cell pairs in the last two rows code the location of the tracked edge stickers 1,3,5,7,17,21,43,47,25,27,29,31, see Table~\ref{tab:STICKER2-edge}. This n-tuple coding requires tuple cells with varying number of position values and leads to 
$$ 8^8\cdot 3^8\cdot 12^{12}\cdot 2^{12} = \mathbf{4.0\cdot 10^{27}}$$
combinations in representation STICKER2.\footnote{
This is, by the way, identical to $(8\cdot 3)^8 \cdot (12\cdot 2)^{12}=24^{(8+12)}=24^{20} = 4.0\cdot 10^{27}$, the same number we had above in the second mode of STICKER. But STICKER2 has the advantage that the combinations are spread over more board cells (40) than in STICKER (20). 
By having more board cells with fewer position values, the n-tuples can better represent the relationships between cube states.}

\begin{table}[h]
\caption{STICKER2 board representation for the default 3x3x3 cube. For the \BoardVector, cells are numbered row-by-row from 0 to 39.}
\label{tab:board-STICKER2-3x3}
\renewcommand{\arraystretch}{1.35}
\centerline{
\begin{tabular}{l|c|c|c|c|c|c|c|c|c|c|c|c|c} \cline{4-10}
\multicolumn{3}{l|}{corner  }& a\re & b\re & c\re & d\re & e\re & f\re & g\re & h\re & \multicolumn{2}{|l}{} & {\scriptsize 8 positions}\\ \cline{4-11}
\multicolumn{3}{l|}{\faceID }  & 1    & 1    & 1    & 1    & 1    & 1    & 1    & 1    & \multicolumn{2}{|l}{} & {\scriptsize 3 positions} \\ \cline{2-13}
edge   & A\re  & B\re & C\re & D\re & E\re & F\re & G\re & H\re & I\re & J\re & K\re & L\re & {\scriptsize 12 positions} \\ \cline{2-13}
face ID& 1     & 1    & 1    & 1    & 1    & 1    & 1     & 1   & 1    & 1    & 1    & 1    & {\scriptsize 2 positions} \\ \cline{2-13}
\end{tabular}
}
\renewcommand{\arraystretch}{1.0}
\end{table}

\subsection{Adjacency Sets}
\label{sec:adjacency3x3}

To create n-tuples by random walk, we need to define adjacency sets (sets of neighbors) for every board cell $k$. 

For CUBESTATE, the board is the flattened representation of the 3x3x3 cube (Fig.~\ref{fig:3x3x3stickers}). The adjacency set is defined as the 4-point neighborhood, where two stickers are neighbors if they are neighbors (share a common edge) on the cube. 

For STICKER2, the board consists of 40 cells shown in Table~\ref{tab:board-STICKER2-3x3}. Since it matters for the corner stickers mostly where the other corner stickers are and for the edge stickers mostly where the other edge stickers are, it is reasonable to form two adjacency subsets $S_1=\{00,\ldots,15\}$ and $S_2=\{16,\ldots,39\}$ and to define the adjacency set 
				$$\mbox{Adj}(k)=S_i\, \backslash\, \{k\}$$
for each $k \in S_i,\, i=1,2$.

\section{Hyperparameters}
\label{app:hyperparams}
In this appendix we list all parameter settings for the GBG agents used in this paper. Parameters were manually tuned with two goals in mind: (a) to reach high-quality results and (b) to reach stable (robust) performance when conducting multiple training runs with different random seeds. The agents listed further down are the best-so-far agents found (best among all agents that learn from scratch by self-play). 

The detailed meaning of RL parameters is explained in \cite{Konen2021_FARL_arXiv}: 
\begin{itemize}
	\item Algorithms 2, 5 and 7 in \cite{Konen2021_FARL_arXiv} explain parameters $\alpha$ (learning rate), $\gamma$ (discount factor), $\epsilon$ (exploration rate) and output sigmoid $\sigma$ (either identity or $\tanh$).
	\item Appendix A.3 explains our eligibility method, parameters are: eligibility trace factor $\lambda$, horizon cut $c_h$, eligibility trace type ET (normal) or RESET (reset on random move). If not otherwise stated, we use in this paper $\lambda=0$ (no eligibility traces). For $\lambda=0$, horizon cut $c_h$ and eligibity trace type are irrelevant. If $\lambda>0$, their defaults $c_h=0.1$ and trace type ET apply.
	\item Appendix A.5 explains our TCL method (also summarized in Sec.~\ref{sec:tcl}). Parameters of TCL are: TC-Init (initialization constant for counters), TC transfer function (TC-id or TC-EXP),  $\beta$ (exponential factor in case of TC-EXP), TC accumulation type (delta or recommended weight-change). 
\end{itemize} 

Another branch of our algorithm is the MCTS wrapper, which can be used to wrap TD-N-tuple agents during evaluation and testing. MCTS wrapping is briefly explained in Sec.~\ref{sec:mcts}. The precise algorithm for MCTS wrapping is explained in detail in \cite[Sec. II-B]{Scheier2022}.\footnote{As \cite[Sec. IV-E]{Scheier2022} shows, the MCTS wrapper may be used as well during training, but due to large computation times needed for this, we do not follow that route in this paper.} Parameters of MCTS are:
\begin{itemize}
	\item $c_{PUCT}$: relative weight for the prior probabilities of the wrapped agent in relation to the value that the wrapper estimates
	\item $d_{max}$: maximum depth of the MCTS tree, if -1: no maximum depth
	\item UseSoftMax: boolean, whether to use SoftMax normalization for the priors or not
	\item UseLastMCTS: boolean, whether to re-use the MCTS from the previous move within an episode or not
\end{itemize}

\noindent
Further parameter explanations: 
\begin{itemize}
	\item Sec.~\ref{sec:represent-ntuple} in this document explains n-tuples, parameters are: number of n-tuples, length of n-tuples, and n-tuple creation mode (fixed, random walk, random points). 
	\item Sec.~\ref{sec:symmetr} in this document explains symmetries. If parameter $\texttt{nSym}=0$, do not use symmetries. If $\texttt{nSym}>0$, use this number \texttt{nSym} of symmetries. In the Rubik's cube case, \texttt{nSym} is a number between 0 and 24. 
	\item LearnFromRM: whether to learn from random moves or not. (Does not apply here, because we use in Rubiks's cube always $\epsilon=0$, i.e. we have no random moves.)
	\item ChooseStart-01: whether to start episodes from different 1-ply start states or always from the default start state. (Does not apply here, because we start in Rubik's cube never from the default cube, but always from the $p$-twisted cube.)
	\item $E_{train}$: maximum episode length during training, if -1: no maximum length.
	\item $E_{eval}$: maximum episode length during evaluation and play, if -1: no maximum length.
\end{itemize} 

All agents were trained with no MCTS wrapper inside the training loop. 
The hyperparameters of the agent for each cube variant were found by manual fine-tuning.
See also~\citep{Konen22a}. 

In the following, we list the precise settings for all agents used in this paper. If not stated otherwise, these common settings apply to all agents: sigmoid $\sigma = \mbox{\textit{id}}$, LearnFromRM = false, ChooseStart-01 = false. Wrapper settings during test and evaluation: MCTS wrapper with $c_{PUCT}=1.0$, $d_{max}=50$, UseSoftMax = true, UseLastMCTS = true. 

Common parameters of Algorithm~\ref{algo:TDNTuple4-rubiks} in Sec.~\ref{sec:tdntuple4} are: cost-to-go $c=-0.1$ and positive reward $R_{pos}=1.0$.

\vspace{0.5cm}
The parameters for training without symmetries ($\texttt{nSym}=0$) in Sec.~\ref{sec:resWrapper} are:
\begin{itemize}
\item \textbf{2x2x2 cube, \HTM}: $\alpha=0.25$, $\gamma=1.0$, $\epsilon=0.0$, $\lambda=0.0$, no output sigmoid. N-tuples: 60 7-tuples created by random walk. TCL activated with  transfer function TC-id, TC-Init$=10^{-4}$ and rec-weight-change accumulation. 3,000,000 training episodes. $p_{max}=13$, $E_{train}=16$, $E_{eval}=50$.\\
	{\scriptsize Agent filename in GBG: 2x2x2\_STICKER2\_AT/TCL4-p13-ET16-3000k-60-7t-stub.agt.zip}
\item \textbf{2x2x2 cube, \QTM}: same as \textit{2x2x2 cube, \HTM}, but with $p_{max}=16$, $E_{train}=20$.\\
	{\scriptsize Agent filename in GBG: 2x2x2\_STICKER2\_QT/TCL4-p16-ET20-3000k-60-7t-stub.agt.zip}
\item \textbf{3x3x3 cube, \HTM}: same as \textit{2x2x2 cube, \HTM}, but with  120 7-tuples created by random walk, $p_{max}=9$, $E_{train}=13$.  \\
	{\scriptsize Agent filename in GBG: 3x3x3\_STICKER2\_AT/TCL4-p9-ET13-3000k-120-7t-stub.agt.zip}
\item \textbf{3x3x3 cube, \QTM}: same as \textit{3x3x3 cube, \HTM}, but with $p_{max}=13$, $E_{train}=16$.\\
	{\scriptsize Agent filename in GBG: 3x3x3\_STICKER2\_QT/TCL4-p13-ET16-3000k-120-7t-stub.agt.zip}
\end{itemize}

The agent files given in the list above are just stubs, i.e. agents that are initialized with the correct parameters but not yet trained. This is because a trained agent can require up to 80 MB disk space, which is too much for GitHub. Instead, a user of GBG may load such a stub agent, train it (takes between 10-40 minutes) and save it to local disk.

When evaluating in Sec.~\ref{sec:resWrapper} the trained agents with different MCTS wrappers, we test in each case whether $c_{PUCT}=1.0$ or $10$ is better. In most cases, $c_{PUCT}=1.0$ is better, but for (2x2x2, QTM, 800 iterations) and for (3x3x3, HTM, 100 iterations) $c_{PUCT}=10.0$ is the better choice.

\vspace{0.5cm}
The parameters for training with symmetries ($\texttt{nSym}=8, 16, 24$) in Sec.~\ref{sec:resSymmetry} are:

\begin{itemize}
\item \textbf{3x3x3 cube, \QTM}: same as \textit{3x3x3 cube, \QTM} in Sec.~\ref{sec:resWrapper}, but with $n_{sym}=8, 16,24$.\\
	{\scriptsize Agent filename in GBG: 3x3x3\_STICKER2\_QT/TCL4-p13-ET16-3000k-120-7t-nsym08-stub.agt.zip,\\
	 \hspace*{2.8cm} 3x3x3\_STICKER2\_QT/TCL4-p13-ET16-3000k-120-7t-nsym16-stub.agt.zip,\\
	 \hspace*{2.8cm} 3x3x3\_STICKER2\_QT/TCL4-p13-ET16-3000k-120-7t-nsym24-stub.agt.zip.\\}
\end{itemize}

Again, the agent filenames are just stubs, i.e. agents that are initialized with the correct parameters but not yet trained. As above, a user of GBG may load such a stub agent, train it (which takes in the symmetry case between 5.4h and 13h, see Table~\ref{tab:compTimes}) and save it to local disk.

For further details and experiment shell scripts, see also the associated \textbf{Papers-with-Code repository} \href{https://github.com/WolfgangKonen/PapersWithCodeRubiks}{\url{https://github.com/WolfgangKonen/PapersWithCodeRubiks}}.

\end{document}